\documentclass{article}
\usepackage{microtype}
\usepackage{graphicx}
\usepackage{subfigure}
\usepackage{booktabs} % for professional tables
\usepackage{hyperref}       % hyperlinks
\usepackage{url}            % simple URL typesetting
\usepackage{amsfonts}       % blackboard math symbols
\usepackage{nicefrac}       % compact symbols for 1/2, etc.
\usepackage{lipsum}
\usepackage{fancyhdr}       % header
\graphicspath{{media/}}     % organize your images and other figures under media/ folder
\usepackage{xspace}
\usepackage{amsmath}
\usepackage{multirow}
\usepackage{fontawesome}
\usepackage{color}
\usepackage{pifont}
\usepackage{bm}
\usepackage{fancyhdr}       % header
\usepackage{makecell}      % for \makecell in tables
\usepackage{float}
\usepackage{xcolor}
\usepackage{tikz}
\newcommand{\nickname}{\textit{db}-SP\xspace}

\definecolor{purple}{rgb}{0.5,0,0.8}
\usepackage[accepted]{mlsys2025}

\mlsystitlerunning{Accelerating Sparse Attention for Visual Generative Models with Dual-Balanced Sequence Parallelism}

\begin{document}

\twocolumn[
\mlsystitle{\textit{db}-SP: Accelerating Sparse Attention for Visual Generative Models with Dual-Balanced Sequence Parallelism}

% DualSP: Accelerating Sparse Attention for Visual Generation Models with Dual-Balanced Sequence Parallelism
% DualSP: Accelerating Sparse Attention for Visual Generation Models with Dual-Balanced Workload Grouping
% Striped Attention: Ring Attention workload partition

% unless you've provided the [accepted] option to the mlsys2025


\mlsyssetsymbol{equal}{*}

\begin{mlsysauthorlist}
\mlsysauthor{Siqi Chen}{equal,thu}
\mlsysauthor{Ke Hong}{equal,thu}
\mlsysauthor{Tianchen Zhao}{thu}
\mlsysauthor{Ruiqi Xie}{thu}
\mlsysauthor{Zhenhua Zhu}{thu}
\mlsysauthor{Xudong Zhang}{thu}
\mlsysauthor{Yu Wang}{thu}
\end{mlsysauthorlist}

\mlsysaffiliation{thu}{Department of Electronic Engineering, Tsinghua University, Beijing, China}

\mlsyscorrespondingauthor{Yu Wang}{yu-wang@mail.tsinghua.edu.cn}

\mlsyskeywords{Machine Learning, MLSys}

\vskip 0.3in

\begin{abstract}
% This document provides a basic paper template and submission guidelines.
% Abstracts must be a single paragraph, ideally between 4--6 sentences long.
% Gross violations will trigger corrections at the camera-ready phase.
Scaling Diffusion Transformer (DiT) inference via sequence parallelism is critical for reducing latency in visual generation, but is severely hampered by workload imbalance when applied to models employing block-wise sparse attention. The imbalance stems from the inherent variation in sparsity across attention heads and the irregular distribution of dense blocks within the sparse mask, when sequence parallelism is applied along the head dimension (as in Ulysses) or the block dimension (as in Ring Attention). In this paper, we formalize a \textit{sparse imbalance ratio} to quantify the imbalance, and propose \nickname, a sparsity-aware sequence parallelism technique that tackles the challenge. \nickname contains a dual-level partitioning approach that achieves near-perfect workload balance at both the head and block levels with negligible overhead. Furthermore, to handle the evolving sparsity patterns across denoising steps and layers, \nickname dynamically determines the parallel degrees for the head and block dimensions at runtime. Experimental results demonstrate that \nickname delivers an end-to-end speedup of 1.25× and an attention-specific speedup of 1.40× over state-of-the-art sequence parallel methods on average. Code is available at https://github.com/thu-nics/db-SP.
\end{abstract}
]
\begin{tikzpicture}[remember picture, overlay]
    % xshift 和 yshift 可以用来微调徽章在右上角的位置
    \node[anchor=north east, xshift=-1.5cm, yshift=-0.2cm] at (current page.north east) {
        \href{https://www.acm.org/publications/policies/artifact-review-and-badging-current}{\includegraphics[width=2cm]{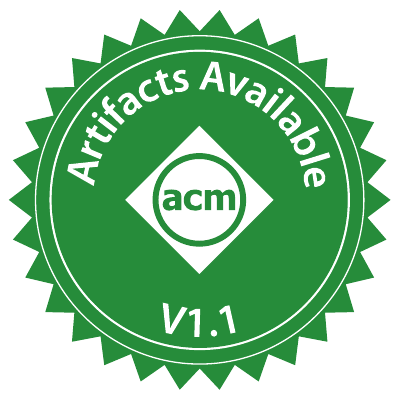}}
        \hspace{0.2cm} % 两个徽章之间的间距
        \href{https://www.acm.org/publications/policies/artifact-review-and-badging-current}{\includegraphics[width=2cm]{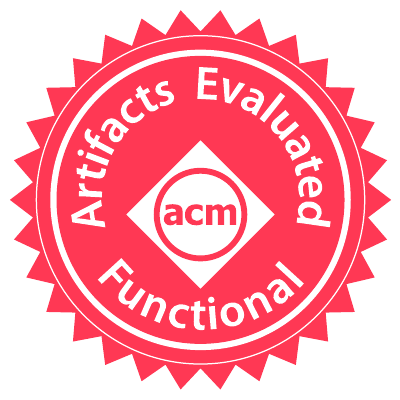}}
        \hspace{0.2cm} % 两个徽章之间的间距
        \href{https://www.acm.org/publications/policies/artifact-review-and-badging-current}{\includegraphics[width=2cm]{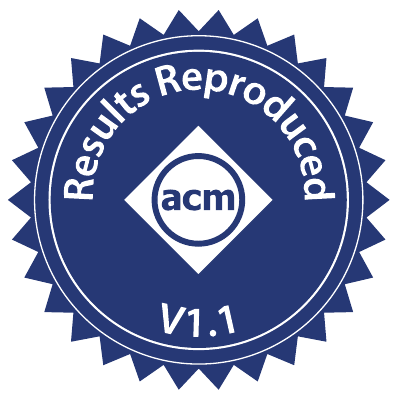}}
    };
\end{tikzpicture}



\printAffiliationsAndNotice{\mlsysEqualContribution} % otherwise use the standard text.

\section{Introduction}\label{intro}
% \hk{[+ab ~2 pages.]}

% DiT在视觉生成任务中的重要性；DiT的计算流程，Transforomer layer中主要是attention和线性层，其中attention时间占比很大
Recent advances in visual generative models, particularly those based on Diffusion Transformers (DiTs), have revolutionized image and video generation ability. The computational flow of a DiT involves feeding a noised latent and conditioning information into a Transformer~\cite{vaswani2023attentionneed} backbone, which predicts the noise to denoise the latent over multiple denoising steps. A Transformer layer is primarily composed of attention and linear computations. Notably, the attention computation suffers from quadratic computational complexity with respect to the token number, and hence becomes a significant bottleneck in DiT inference, occupying over half of the end-to-end latency in multi-frame video generation tasks. 
% These models leverage the scalable architecture of Transformers to achieve state-of-the-art results in high-resolution image and long-video generation tasks. However, the core component of the Transformer—the self-attention mechanism—suffers from quadratic computational complexity with respect to the sequence length. This becomes a significant bottleneck when processing the lengthy token sequences common in visual data, such as high-resolution images or multi-frame videos.

\begin{figure}[t]
    \centering
    \includegraphics[width=0.99\linewidth]{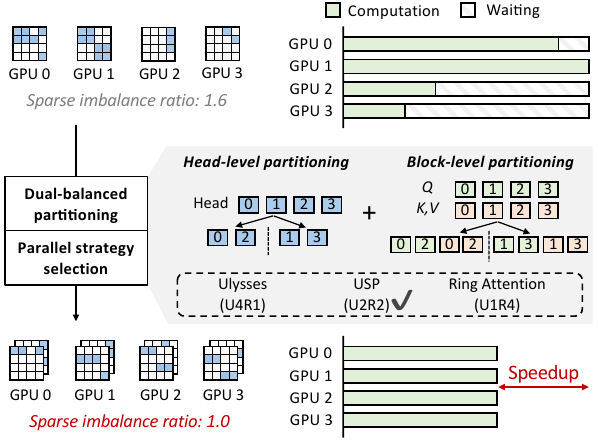}
    \vspace{-0.em}
    \caption{\nickname achieves speedup via dual-balanced partitioning and parallel strategy selection. \textit{Top}: Sparse attention is partitioned across GPUs by attention heads, leading to workload imbalance. \textit{Bottom}: \nickname balances the workload at both head-level and block-level partitioning. Ulysses, USP, and Ring Attention are different sequence parallelism methods, where U$x$R$y$ denotes that $x$ GPUs perform Ulysses and $y$ GPUs perform Ring Attention. } 
    \vspace{-2.0em}
    \label{fig:intro}
\end{figure}

% 稀疏attention优化对于降低attention延时的作用，其中block-wise sparse是能够在保证质量基础上，对硬件更加友好，本文主要关注block-wise sparse attention。但是即使使用了稀疏，单卡上的latency仍然很长。介绍sequence parallel对于DiT推理的重要作用，基于计算密集型的特性，SP能够几乎线性降低延时。介绍主要的SP方法。
% To mitigate this, block-wise sparse attention has emerged as a promising approach. Exploiting the inherent sparsity in attention maps, it reduces computation while maintaining generation quality. Meanwhile, sequence parallelism (SP)—including paradigms like Ulysses and Ring Attention—has been adopted to distribute long sequences across multiple devices, overcoming the memory and computational limits of single GPUs. While both techniques offer individual benefits, their combination appears natural for extreme-scale visual generation.
Sparse attention~\cite{zhang2025spargeattn, xi2025sparse, yang2025sparse, xia2025adaspa, zhao2025paroattentionpatternawarereorderingefficient, ribar2024sparqattentionbandwidthefficientllm, yuan2024ditfastattnattentioncompressiondiffusion, fu2025slidingwindowattentiontraining} is an effective technique to enhance attention efficiency while maintaining generation quality. Among them, the block-wise sparse pattern~\cite{zhao2025paroattentionpatternawarereorderingefficient, zhang2025spargeattn, yang2025sparse, xia2025adaspa} becomes promising and is widely adopted in visual generation models due to its efficiency advantages. This efficiency stems from its ability to maintain local computational contiguity and assign adequately sized data to distinct Streaming Multiprocessors (SMs), thereby improving hardware utilization. Nevertheless, even with block-wise sparse attention, the generation latency on a single GPU remains prohibitively high, \textit{e.g.}, exceeding 15 minutes per video using the Wan2.1-T2V-14B model on an A800 GPU with PAROAttention, which is still impractical for real-world use.

Sequence parallelism (SP) plays a critical role in scaling DiT inference. Given that DiT models typically process a large number of tokens but have relatively small model weights, partitioning the activations is more critical than partitioning the weights. Moreover, the compute-intensive nature ensures that partitioning the activations across multiple GPUs still provides each with sufficient workload to saturate the computational capabilities. Consequently, by applying SP, the latency of DiT inference can be scaled down almost proportionally across multiple GPUs. 

% 然而，现有的SP方法都忽略稀疏带来的负载问题，存在严重的负载不均衡。分别分析负载不均衡的来源。
% However, naively integrating sparse attention with existing sequence parallelism strategies leads to severe workload imbalance across devices. This imbalance arises from two distinct sources: 1) Head-level imbalance, where different attention heads exhibit varying sparsity patterns, and 2) Token-level imbalance, where the distribution of non-zero blocks in the attention mask is non-uniform across token chunks. As a result, the potential acceleration from sparsity is undermined by synchronization delays, as the slowest device dictates the pace of the entire system. 
However, existing SP methods, including Ulysses~\cite{jacobs2023deepspeedulyssesoptimizationsenabling}, Ring Attention~\cite{liu2023ringattentionblockwisetransformers}, and USP~\cite{fang2024uspunifiedsequenceparallelism}, overlook the workload imbalance introduced by sparse attention, failing to fully leverage the potential of multi-GPU parallelism. Specifically, Ulysses partitions different attention heads across GPUs, while Ring Attention partitions the sequence (and thus the dense blocks in sparse attention). USP combines the two aforementioned methods by organizing each GPU into a Ulysses group and a Ring Attention group, allowing the respective parallel method to be applied orthogonally within the dedicated group. Therefore, the workload imbalance manifests at two distinct levels. (1) \textbf{Head-level}: The sparsity pattern varies significantly across different attention heads, leading to unequal assignments to each GPU. (2) \textbf{Block-level}: The irregular distribution of dense blocks within the sparse attention mask results in an uneven number of blocks across different GPUs.

% 为了进一步量化分析，我们引入sparse imbalance ratio，xxx。sparse ratio大概在什么范围内，表示具有多少潜在的性能提升上界。
% In this paper, we identify and formalize this problem by introducing the Sparse Imbalance Ratio ($\rho_s$), a metric to quantify workload imbalance under different sparsity patterns and parallelization strategies. Our measurements reveal that $\rho_s$ can reach up to 1.513 in real-world models, indicating substantial room for improvement.
In this paper, we formalize such workload imbalance by introducing a \textbf{\textit{sparse imbalance ratio}} ($\rho_s$), which indicates the ratio of the workload on the most heavily-loaded GPU to the average number across all GPUs. Our measurements reveal that the \textit{sparse imbalance ratio} achieves over 1.5 in visual generation tasks (1.513 with Wan2.1-T2V-14B model and SpargeAttn using Ulysses on eight A800 GPUs), indicating substantial potential for improvement. 

% 但是，在实际系统中，想要minimize这个ratio，存在以下挑战：xxx。
Although mitigating the workload imbalance potentially brings acceleration, there remain several challenges. (1) \textbf{Dual-level interplay}: The two levels of workload imbalance are coupled, making the co-optimization a non-trivial challenge that often results in prohibitively intricate partitioning plans. (2) \textbf{Reorganizing overhead}: Workload balancing requires data replacement, leading to intra-GPU reorderings and inter-GPU exchanges. The challenge is to mitigate the overhead for higher performance gain. (3) \textbf{Parallel strategy selection}: Selecting an SP strategy (\textit{i.e.}, the parallel degrees for Ulysses and Ring Attention) presents a challenge when integrated with sparse attention, as the workload balance is highly dependent on both the sparse pattern and the parallel strategy.

% 为此，我们做了什么。
To address the challenges, we propose \nickname, a dual-balanced sequence parallelism that balances the workload of sparse attention across GPUs at both levels. At both levels, \nickname employs a greedy algorithm for workload partitioning. To reduce overhead, \nickname exploits sparse mask similarity across denoising steps to reduce partitioning executions, and introduces a reward factor at the block level to create a biased greedy algorithm that mitigates inter-GPU exchange costs. Given the interplay between the two levels, our design is guided by a key prior: near-perfect load balance is achievable independently at each level. Consequently, \nickname simplifies the joint optimization problem into two sequential sub-problems: it first performs head-level partitioning, then conducts block-level partitioning under the assumption of an ideally balanced head-level workload distribution. Furthermore, \nickname incorporates a runtime mechanism that dynamically switches between parallel strategies through latency prediction to select the optimal strategy and the corresponding partitioning plan.

The proposed approach is straightforward yet effective, spanning multi-level techniques. (1) \textbf{Kernel implementation}:  \nickname utilizes sorting-based partitioning for workload balancing, avoiding the degraded computation-to-memory-access ratio that results from decomposing dense blocks into small tiles. (2) \textbf{Parallelism optimization}: \nickname yields a near-perfectly balanced workload across GPUs and introduces dynamic selection of Ulysses and Ring Attention parallel degrees at runtime, as SP does not affect model weight placement. (3) \textbf{Application characteristics}:  \nickname exploits the temporal similarity and initial data residence to reduce overhead.

In summary, this paper makes the following contributions:
\vspace{-0.8em}
% 总结。
\begin{itemize}
    \item We identify and systematically analyze the dual-level (head and block) workload imbalance that arises from applying block-wise sparse attention with existing sequence parallelism methods.
    \vspace{-0.4em}
    \item We introduce a dual-balanced workload partitioning approach that decouples the joint optimization of the two levels, employing distinct algorithms for the head-level and block-level workload balancing at a negligible overhead.
    \vspace{-0.4em}
    \item We design a sparsity-aware strategy selection mechanism that predicts the latency of different parallelism plans, enabling dynamic selection of the optimal parallel strategy for each attention computation.
    \vspace{-0.4em}
    \item We implement \nickname and experiments shown that \nickname achieves up to 1.25$\times$ end-to-end speedup and 1.40$\times$ attention acceleration on average over state-of-the-art sequence parallel methods on typical DiT models and tasks.
    \vspace{-0.4em}
\end{itemize}
% We identify and systematically analyze the dual-level (head and token) workload imbalance problem that arises when combining block-wise sparse attention with existing sequence parallelism methods.

% We introduce a complete set of grouping algorithms for both head-level and token-level workload balancing, which can be applied independently or jointly according to the parallelism settings.

% We design a sparsity-aware strategy selector that models the expected runtime of different parallelism plans, enabling adaptive selection of the optimal parallelization strategy per layer and denoising step.

% We implement db-SP and demonstrate that it achieves up to 1.25$\times$ end-to-end speedup and 1.5$\times$ attention acceleration on large-scale models like Wan2.1 and CogVideoX, with minimal overhead.

% By enabling the efficient synergy of sparse attention and sequence parallelism, db-SP paves the way for faster and more scalable visual generation models.
\section{Background}\label{sec:background}
In this section, we introduce the background of Diffusion Transformer inference in visual generative tasks, and the related works. 

\subsection{Visual Generative Models}
Diffusion Transformer (DiT)~\cite{peebles2023scalable} has emerged as the mainstream model architecture for visual generative tasks such as image and video generation~\cite{wan2025wanopenadvancedlargescale, yang2025cogvideoxtexttovideodiffusionmodels}, incorporating the core denoising steps of diffusion models and the scalable and powerful Transformer~\cite{vaswani2023attentionneed, dosovitskiy2021imageworth16x16words} architecture. The core of the Transformer architecture is the attention mechanism. Taking the Multi-Head Attention (MHA) in DiT models as an example, the query ($Q$), key ($K$), and value ($V$) matrices derived through linear projection are divided into multiple attention heads along the feature dimension, and then the output ($O$) is calculated by $O = PV, P =\text{softmax}(QK^{T}/\sqrt{d})$ for each attention head, where $d$ is head dimension size.

% The Diffusion Transformer (DiT)~\cite{peebles2023scalable} represents a significant architectural evolution in the field of generative AI, particularly for image and video synthesis. It reserves the core denoising process of diffusion models and replaces the commonly used U-Net backbone with a more scalable and powerful Transformer architecture\cite{vaswani2023attentionneed, dosovitskiy2021imageworth16x16words}. This shift leverages the proven strengths of Transformers in handling long-range dependencies and complex conditional inputs, leading to state-of-the-art performance in generative tasks.

% Early video DiT models, such as those in OpenSORA\cite{opensora,opensora2}, often employed a factorized "spatial-temporal" attention mechanism. This approach typically involved applying self-attention within spatial dimensions (height and width) and temporal dimensions (frame axis) separately or sequentially to manage computational complexity. However, recent state-of-the-art models like CogVideoX\cite{yang2025cogvideoxtexttovideodiffusionmodels} and Wan\cite{wan2025wanopenadvancedlargescale} have moved towards a more integrated "3D full attention" mechanism\cite{kong2025hunyuanvideosystematicframeworklarge, arnab2021vivitvideovisiontransformer}. This paradigm treats a video clip as a unified sequence of spatial-temporal tokens, allowing the model to jointly attend to visual features across both space and time simultaneously.  This increased model size and token length pose significant challenges for efficient deployment.

Early DiT models in visual generative tasks, such as Open-Sora series~\cite{opensora,opensora2}, employ a factorized spatial-temporal attention mechanism, computing self-attention along the spatial dimensions (\textit{i.e.,} height and width) and the temporal dimension (\textit{i.e.,} frame) independently. Nevertheless, recent progress including CogVideoX~\cite{yang2025cogvideoxtexttovideodiffusionmodels} and Wan~\cite{wan2025wanopenadvancedlargescale} utilizes a more integrated 3-dimensional self-attention~\cite{kong2025hunyuanvideosystematicframeworklarge, arnab2021vivitvideovisiontransformer} that treats a video clip as a unified sequence of spatial-temporal tokens, allowing the model to jointly attend to visual features across both space and time simultaneously. Such a mechanism further prolongs the attention latency in the inference procedure, degrading the generation efficiency, particularly for high-resolution image or long-video generation tasks. % \hk{\textit{E.g.}, when xxx, the attention computation latency occupies xxx\% of the whole generation duration.}

\begin{figure}[t]
    \centering
    \includegraphics[width=0.99\linewidth]{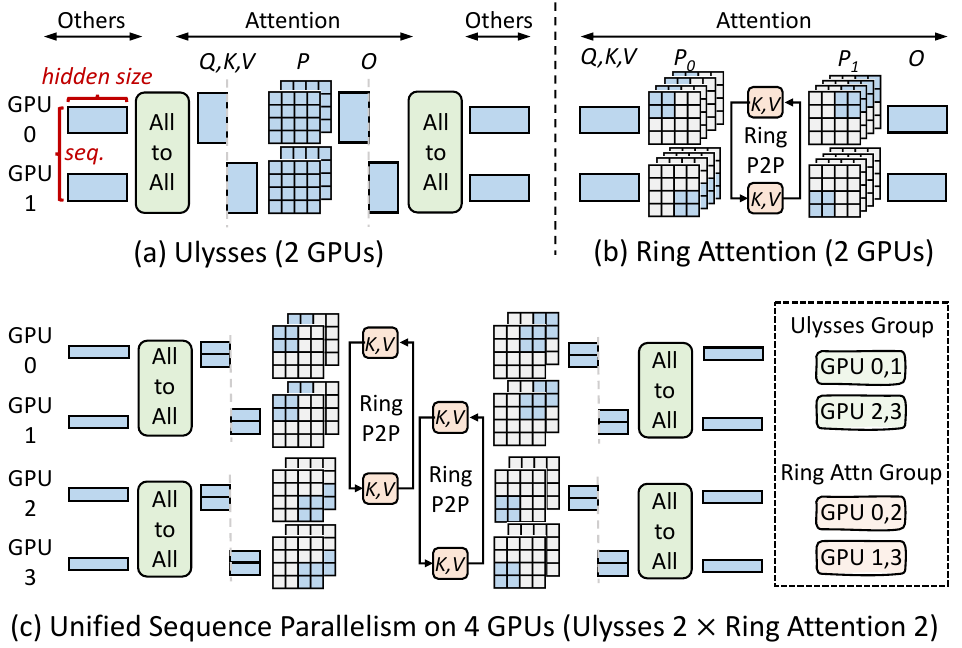}
    \vspace{-1.0em}
    \caption{Comparison of sequence parallelism variants. } 
    \vspace{-1.0em}
    \label{fig:sp}
\end{figure}

\subsection{Block-wise Sparse Attention}
In visual generative models, the attention map $P =\text{softmax}(QK^{T}/\sqrt{d})$ exhibits inherent sparsity, and previous works~\cite{zhao2025paroattentionpatternawarereorderingefficient, zhang2025spargeattn, yang2025sparse, xia2025adaspa}  demonstrate that the block-wise sparse attention reduces computational costs while maintaining acceptable generation quality. The block-wise pattern is hardware-friendly, avoiding irregular memory accesses within blocks and ensuring sufficient data for parallel computation on GPU Streaming Multiprocessors (SMs). 

% Customized GPU kernels have been developed to accelerate block-wise sparse attention~\cite{zhao2025paroattentionpatternawarereorderingefficient, yang2025sparse} on a single GPU. Following FlashAttention2~\cite{dao2023flashattention2}, $Q$ is partitioned along the sequence dimension into $b^{Q}$ chunks. Assume there are $h$ attention heads, and then the attention workload can be divided into $b^{Q} \times h$ partitions and dispatched onto SMs for parallel execution. 
% With full attention, the $b^{Q} \times h$ partitions have equal computational workloads (\textit{i.e.}, equal $K,V$ length). When combined with sparse attention, for the $i$-th $Q$ segment in the $j$-th attention head, its corresponding $K, V$ length is different from others.
% In detail, for each partition, $K$ and $V$ are also segmented into $b^{KV}$ chunks and iteratively read into the shared memory for computation. Hence, the block-wise sparse attention kernel can \textbf{naturally adapt to the partitioning scheme in the full attention}, by setting the chunk size equal to the block size in the sparse mask. 
Following FlashAttention2~\cite{dao2023flashattention2}, $Q$ is partitioned along the sequence dimension into sub-sequences, with the workload corresponding to different sub-sequences and attention heads distributed to different SMs for parallel processing. Consequently, for a block-wise sparse attention kernel, if the block size is an integer multiple of the sub-sequence length, it can reuse the underlying implementation of the dense attention kernel without introducing redundant computation. Therefore, for block-wise sparse attention, \textit{\textbf{the total number of dense blocks in the sparse mask across all heads represents the workload on a single GPU}}.

% Trade-offs exist in  
% In visual generative models, the attention map $P =\text{softmax}(QK^{T}/\sqrt{d})$ exhibits inherent sparsity, as each patch typically attends only to a limited set of other patches. Previous work has proved that this sparsity pattern is generally block-wise. Such block-wise sparsity arises because video content usually has strong local correlations in both time and space, meaning that a patch primarily focuses on information from either the same spatial location across different frames (temporal attention) or within the same frame but different spatial locations (spatial attention). Previous research has leveraged this property to design efficient sparse attention mechanisms, achieving notable success in maintaining model performance while significantly lowering memory and computational costs. \cite{zhang2025efficient}

% High-performance customized sparse attention operators are 

% These mechanisms can be generally divided into two categories: static sparse attention and dynamic sparse attention. The static sparse attention mechanism is a kind of pattern-based method. It predefines sparsity patterns derived from empirical observations and skips the computation of the corresponding position without other overhead during runtime. Dynamic sparse attention constructs sparse patterns at runtime based on some features of the input data.

\subsection{Sequence Parallelism}
Sequence parallelism (SP) shows advantages in DiT inference over the commonly used Tensor Parallelism (TP), as it avoids the time-consuming AllReduce communication primitive. There are two typical paradigms for sequence parallelism, \textit{i.e.}, Ulysses~\cite{jacobs2023deepspeedulyssesoptimizationsenabling} and Ring Attention~\cite{liu2023ringattentionblockwisetransformers}. Based on that, Unified Sequence Parallelism has been proposed to incorporate the two paradigms to form a hybrid one.

% \subsubsection{Ulysses}
As depicted in Fig.~\ref{fig:sp}(a), Ulysses~\cite{jacobs2023deepspeedulyssesoptimizationsenabling} partitions the input along the sequence dimension across multiple GPUs for all the other parts except for the attention computation, and dispatches different attention heads onto different GPUs for the attention computation. Specifically, the inputs are reorganized via a pair of all-to-all communication primitives before and after the attention, respectively. 
% The communication is of $\mathcal{O}(n/m)$, where $n$ and $m$ represent the sequence length and the GPU number. \hk{[Wrong complexity?]}

% DeepSpeed-Ulysses~\cite{jacobs2023deepspeedulyssesoptimizationsenabling} addresses the critical challenge of training Transformers with extremely long sequences by introducing a novel and communication-efficient sequence parallelism strategy. Unlike existing parallel paradigms (data, tensor, pipeline) that scale other model dimensions, Ulysses specifically targets the sequence length. 
% Its core design partitions the input sequence dimension across available GPUs. Before attention computation, the model employs an all-to-all collective communication on partitioned queries, keys, and values. This exchange ensures that each GPU receives the entire sequence, but only for a distinct, non-overlapping subset of attention heads, enabling parallel attention computation across heads. Subsequently, another all-to-all collective is used to gather the results and re-partition the output along the sequence dimension for subsequent layers. This strategy results in a superior communication complexity of O(N/P), where N is the sequence length and P is the number of parallel devices. 

% \subsubsection{Ring Attention}
As depicted in Fig.~\ref{fig:sp}(b), Ring Attention~\cite{liu2023ringattentionblockwisetransformers} partitions the input along the sequence dimension for all the parts, including the attention computation. To achieve that, Ring Attention introduces a ring-based communication pattern, \textit{i.e.}, each GPU holds a $Q$ chunk along the sequence dimension, iteratively computing partial attention over each $K$,$V$ chunk, and then passes the $K$,$V$ chunk to the neighboring GPU via peer-to-peer (P2P) communication. The procedure is repeated until all $K$,$V$ chunks have been traversed on each GPU. Such a pattern enables the overlap between the $K$,$V$ communication and the partial attention computation, yielding an opportunity for near-zero communication overhead. 

% Ring Attention~\cite{liu2023ringattentionblockwisetransformers} partitions the input along the sequence dimension for all the operations, including the attention computation. To achieve that, Ring Attention leverages a block-wise computation of self-attention and feedforward networks. The core of its parallel strategy involves organizing the computing devices into a logical ring. Each device in the ring holds a distinct block of the sequence. The computation then proceeds in a coordinated, rotating fashion: as each device computes the attention between its local query block and its current key-value block, it simultaneously sends its key-value block to the next device in the ring and receives new key-value blocks from the previous device. This design fully overlaps the communication of key-value blocks with the block-wise attention computation. Crucially, as long as the computation time for a block exceeds the communication time, this strategy introduces near-zero communication overhead. This allows the effective context length to scale linearly with the number of devices in the ring, effectively enabling near-infinite context training and inference without resorting to approximations.

% \hk{USP should be mentioned.}

% \subsubsection{Unified Sequence Parallelism}
Unified Sequence Parallelism (USP)~\cite{fang2024uspunifiedsequenceparallelism, gu2024loongtrain} is a hybrid design on top of the two former paradigms, addressing the parallelism degree limitation by the attention head number in Ulysses, and the inefficiency caused by chunked attention computation in Ring Attention. As depicted in Fig.~\ref{fig:sp}(c), USP organizes the GPUs into orthogonal Ulysses and Ring Attention groups. GPUs in the same Ulysses group compute different attention heads, whereas the GPUs in the same Ring Attention group partition $Q$ along the sequence dimension and utilize the ring-based communication to traverse $K,V$. 

% organizes the sequence parallel process group as a 2D grid: SP-Ulysses operates across the grid's rows, utilizing All-to-All collective communication to partition tensors along the attention head dimension, while SP-Ring operates across the columns, employing a ring-based Peer-to-Peer (P2P) communication pattern to shuttle sequence blocks between devices and overlap communication with computation. This unified design leverages the strengths of both foundational methods: SP-Ulysses's efficient use of optimized attention kernels and SP-Ring's flexibility regarding the number of attention heads, while mitigating their respective weaknesses, such as SP-Ulysses's head-count limitation and SP-Ring's computational inefficiency and load-balancing issues in causal attention. The approach allows flexible configuration of the Ulysses and Ring degrees, making it highly adaptable to diverse model architectures and network hardware topologies.

\subsection{Related Works}
% Sparsification and parallel computing are effective techniques for accelerating DiT inference. 
In this section, we discuss the related works including the sparse attention methods and parallel computing designs, as well as the studies attempting to incorporate both. Moreover, we also list the related works involving other efficient DiT inference techniques, which are orthogonal to this work.
% The self-attention mechanism, despite being the core of DiT's representation power, constitutes a major computational bottleneck during inference. To address this, numerous studies have focused on accelerating the attention computation in DiTs, which can be broadly categorized into three paradigms: sparse attention, cache quantization.

\subsubsection{Sparse Attention Methods}
Prior works have explored various attention mask patterns tailored for visual generation, including window-based pattern~\cite{fu2025slidingwindowattentiontraining, yuan2024ditfastattnattentioncompressiondiffusion}, spatial-temporal pattern~\cite{xi2025sparse}, hybrid mask combination~\cite{jiang2024minference10acceleratingprefilling}, and block-wise pattern~\cite{zhao2025paroattentionpatternawarereorderingefficient, zhang2025spargeattn, yang2025sparse}. 
Regarding the block-wise pattern, PAROAttention~\cite{zhao2025paroattentionpatternawarereorderingefficient} abstracts static masks against varying inputs, whereas SpargeAttn~\cite{zhang2025spargeattn} and Sparse VideoGen2~\cite{yang2025sparse} utilize the dynamic mechanism to selectively compute part of the attention map online. The block-wise pattern enables practical single-GPU acceleration via customized kernels.

% Besides those applying a static mask against various inputs, there are also works~\cite{ribar2024sparqattentionbandwidthefficientllm, singhania2024lokilowrankkeysefficient, zhang2025spargeattn, yang2025sparse} that construct the sparse mask in a dynamic way, and online selectively compute part of the attention map. 

% The block-wise sparse attention~\cite{zhao2025paroattentionpatternawarereorderingefficient, zhang2025spargeattn, yang2025sparse, jiang2024minference10acceleratingprefilling, lu2025mobamixtureblockattention, acharya2025star} has demonstrated significant potential for practical deployment, not only for visual generation but also for text generation. Among them, ParoAttention~\cite{zhao2025paroattentionpatternawarereorderingefficient} applies a static mask for video generation, whereas SpargeAttn~\cite{zhang2025spargeattn} and Sparse VideoGen2~\cite{yang2025sparse} utilize the dynamic mechanism. 

% and token clustering masks(e.g., PAROAttention \cite{zhao2025paroattentionpatternawarereorderingefficient}).

% SparQAttn\cite{ribar2024sparqattentionbandwidthefficientllm} and LokiAttn\cite{singhania2024lokilowrankkeysefficient} construct the sparse mask by carrying full attention with reduced dimensionality. SpargeAttn\cite{zhang2025spargeattn} computes a compressed attention map and then selectively computes those pairs where it accumulates a high score in the compressed attention map. Sparse VideoGen2\cite{yang2025sparse} clusters QK into several categories and computes the attention only for the selected cluster top-p pairs. 

\subsubsection{Parallel Computing Designs}
DistriFusion~\cite{li2023distrifusion} and PipeFusion~\cite{fang2024pipefusion} are designed specifically for DiT inference in multi-GPU systems, reusing features to unveil the opportunity for layer-wise parallelism. Nevertheless, the feature reusing brings uncertainty to the generation quality. 

Sequence parallelism, including Ulysses~\cite{jacobs2023deepspeedulyssesoptimizationsenabling}, Ring Attention~\cite{liu2023ringattentionblockwisetransformers}, and USP~\cite{fang2024uspunifiedsequenceparallelism} is the commonly used parallel computing design for accelerating DiT inference, and has been implemented in the open-source inference engines, such as xDiT~\cite{fang2024xdit} and ParaAttention~\cite{wavespeed2025paraattention}. However, when combined with the sparse attention, existing sequence parallelism paradigms suffer from severe workload imbalance across GPUs. 

BurstAttention~\cite{sun2024burstattentionefficientdistributedattention, sun2025burstengineefficientdistributedframework} attempts to mitigate the imbalance in Ring Attention with block-wise sparsity, by uniformly partitioning a block onto different GPUs to achieve balance. We observe that such a method is not suitable for visual generation, as the sparse block size is necessarily small to retain generation quality. Such a small block size leads to the degradation of the attention kernel performance after partitioning, which is detailed in \S\ref{sec:e2e}.

\subsubsection{Efficient DiT Inference}
Quantization reduces the memory and computational amounts using low-bit precision of model weights or activations, \textit{e.g.}, SageAttention\cite{zhang2025sageattention, zhang2024sageattention2} has explored quantizing $QK^T$ to INT4/8 while employing FP8 (8-bit floating-point) for $PV$. Caching is the technique that exploits the temporal redundancy across denoising steps, reusing intermediate features~\cite{zou2025acceleratingdiffusiontransformerstokenwise, zou2024acceleratingdiffusiontransformersdual} or complete outputs~\cite{liu2025smoothcacheuniversalinferenceacceleration, lv2025fastercachetrainingfreevideodiffusion} for less computation. Besides, DiffServe~\cite{ahmad2025diffserve} constructs model cascades for routing different requests based on the demand, leading to lower latency violation rates. DDiT~\cite{huang2025ddit} proposes a dynamic resource allocation mechanism for efficiently deploying DiT and encoder modules. Those methods are orthogonal to this work.

\section{Motivation}\label{sec:motivation}
% \hk{[~1 pages.]}
In this section, we first clarify the motivation for using sequence parallelism together with block-wise sparse attention, and then define a theoretical \textit{sparse imbalance ratio} to demonstrate the consequential workload imbalance problem. Finally, we pose the challenges in solving the workload imbalance issue in the real system.

\subsection{Sparse Attention with Sequence Parallelism}
\begin{figure}[t]
    \centering
    \includegraphics[width=0.99\linewidth]{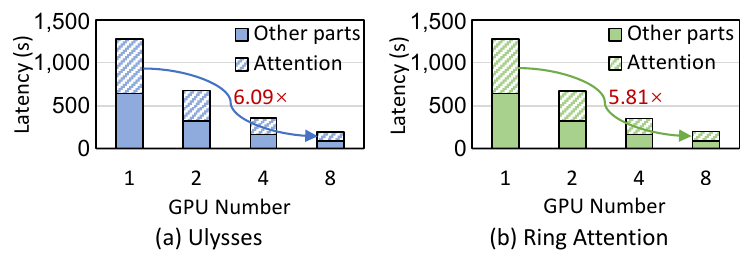}
    \vspace{-1.em}
    \caption{The latency of Wan2.1-T2V-14B with PAROAttention as the GPU number increases. } 
    \vspace{-0.5em}
    \label{fig:scale}
\end{figure}

\textit{Sparse attention is still the primary bottleneck, and applying sequence parallelism is rewarding.}  The self-attention computation in DiT is computationally expensive, growing quadratically with the input token number. As depicted in Fig.~\ref{fig:scale}, when applied with the block-wise sparsity, the sparse attention still occupies over 50\% of the total inference latency. The major components in DiT, including the linear layers and self-attention, are computation-intensive operations with large enough input size ($>$70k tokens), and thus are scalable with sequence parallelism. As depicted in Fig.~\ref{fig:scale}, the end-to-end latency decreases with more GPUs. However, due to workload imbalance, existing sequence parallelism methods fall short of fully realizing their scaling potential, achieving only a 6.09$\times$ (Ulysses) and 5.81$\times$ (Ring Attention) reduction in latency when scaling from single-GPU inference to eight GPUs.

\begin{figure}[t]
    \centering
    \includegraphics[width=0.99\linewidth]{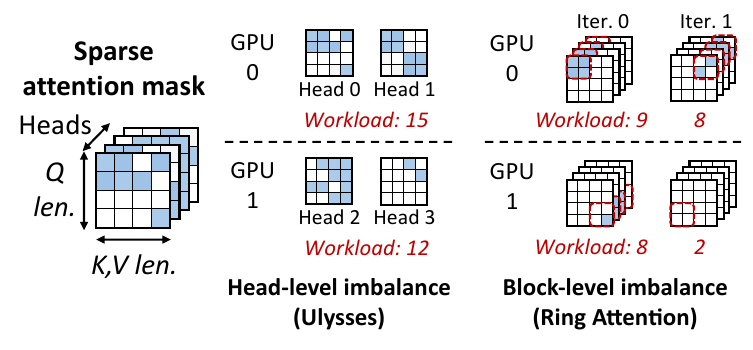}
    \vspace{-1.5em}
    \caption{Dual-level workload imbalance when applying sequence parallelism to sparse attention. } 
    \vspace{-1.0em}
    \label{fig:imbalance}
\end{figure}

% 为什么并行稀疏 存在问题 能优化 数学？
% \textbf{Attention operation in DiTs is the primary bottleneck}. While the self-attention mechanism is fundamental to Diffusion Transformers (DiTs), it is computationally expensive, scaling quadratically ($O(n^2)$) with the input token length. This inherent cost motivates the exploration of various acceleration techniques as we have discussed in Sec.2.

% Parallelism leverages multi-GPU architectures to distribute the computational workload, thereby increasing throughput without precision loss. Among other techniques that has precision loss, sparse attention offers a larger optimization space by enabling the design of various sparse patterns. This significant potential has made it one of the most extensively researched methods in the field today. So in order to accelerate the inference speed and adapt to different kinds of settings and longer token length, combining sparse attention with parallelism is necessary and rewarding. 

\textit{Directly applying sequence parallelism to sparse attention leads to workload imbalance across GPUs.} The workload imbalance arises from two perspectives: (1) Head-level imbalance. Each attention head owns a single attention mask, and the sparsity distinguishes significantly among different heads. When applied with Ulysses, heads are distributed across GPUs, and thus the sparsity among GPUs is different. (2) Block-level imbalance. The sparse attention mask is inherently irregular rather than uniform. When applied with Ring Attention, each GPU holds one $Q$ chunk, meanwhile iteratively computing the corresponding $K, V$ chunks. At every iteration, the dense block number within the $K,V$ chunk for different $Q$ chunk distinguishes, thereby leading to workload imbalance among GPUs.

\subsection{Sparse Imbalance Ratio ($\rho_s$)} 

We introduce a \textit{sparse imbalance ratio} $\rho_s$ to quantify such workload imbalance. Workload imbalance brings latency degradation due to synchronization across GPUs. Specifically, Ulysses synchronizes GPUs in the All-to-All communications after attention, and Ring Attention requires synchronization at each iteration via Ring P2P communication. Therefore, we consider the synchronization in the formulation by decomposing the layer-wise latency into $N$ periods based on the synchronization points. Suppose the GPU set is $\mathcal{G}$, $\rho_s$ is defined as follows.
\vspace{-0.5em}
\begin{align}\label{eq:ratio}
\rho_s = \frac{\sum_{i\in \{0, ..,N-1\}}(\max_{g\in\mathcal{G}}(b_{i, g}({\bf{s,p}})))}{\sum_{i\in \{0, ..,N-1\},g\in\mathcal{G}} b_{i,g}({\bf{s,p}}) / |\mathcal{G}|},
\end{align}
\vspace{-1.5em}

where $b_{i,g}(\bf{s,p})$ denotes the dense block number in the attention masks on GPU $g$ during the $i$-th synchronization period, which is a function of the sparse pattern $\bf{s}$ and parallelism strategy $\bf{p}$. Thus, $\rho_s$ is defined as the ratio of the layer-wise latency given $\bf{s}$ and $\bf{p}$ to the ideal latency under a perfectly balanced workload, \textit{i.e.}, the theoretical speedup through workload balancing. 

As shown in Tab.~\ref{imbalance_ratio}, we measure the sparsity imbalance ratio $\rho_s$ in real-world scenarios with the sparse attention method PAROAttention~\cite{zhao2025paroattentionpatternawarereorderingefficient} and SpargeAttn~\cite{zhang2025spargeattn}. $\rho_s$ ranges from 1.159 to 1.513, indicating potentials for acceleration through workload balancing.

\begin{table}[t]
\caption{The sparse imbalance ratio ($\rho_s$) of different models across various sparse attention methods and sequence parallelism paradigms. A larger $\rho^{\text{s}}$ represents a more imbalanced workload. USP (U$x$R$y$) denotes $x$ GPUs to form a Ulysses group and $y$ GPUs for a Ring Attention group. SpargeAttn has not been implemented with Ring Attention. }
\vspace{-1.8em}
\label{imbalance_ratio}
\vskip 0.15in
\begin{center}
\begin{small}
% \begin{sc}
\begin{tabular}{lccc}
\toprule
Model & \makecell{Sparse Attention\\Method} &  \makecell{Sequence\\Parallelism} & \makecell{$\rho_s$} \\
\midrule
\multirow{5}{*}{Wan2.1} 
 & \multirow{4}{*}{PAROAttention} 
 & Ulysses & 1.355 \\
 & & Ring Attention & 1.255 \\
 & & USP (U4R2) & 1.253 \\
 & & USP (U2R4) & 1.246 \\
 \cmidrule(lr){2-4}
 & \multirow{1}{*}{SpargeAttn} 
 & Ulysses & 1.428 \\
\midrule
\multirow{5}{*}{CogVideoX1.5}
 & \multirow{4}{*}{PAROAttention} 
 & Ulysses & 1.447 \\
 & & Ring Attention & 1.269 \\
 & & USP (U4R2) & 1.201 \\
 & & USP (U2R4) & 1.159 \\
 \cmidrule(lr){2-4}
 & SpargeAttn  & Ulysses & 1.513 \\
\bottomrule
\end{tabular}
% \end{sc}
\end{small}
\end{center}
\vskip -0.2in
\end{table}

% As our method balances the workload from dual perspective (head-level and token-level), it is proved mathematically that for any given sparsity pattern, as long as the sparsity is not too low, it can be made close enough to workload-balanced through head-level or token-level partitioning.

\begin{figure*}[t]
    \centering
    \includegraphics[width=\textwidth]{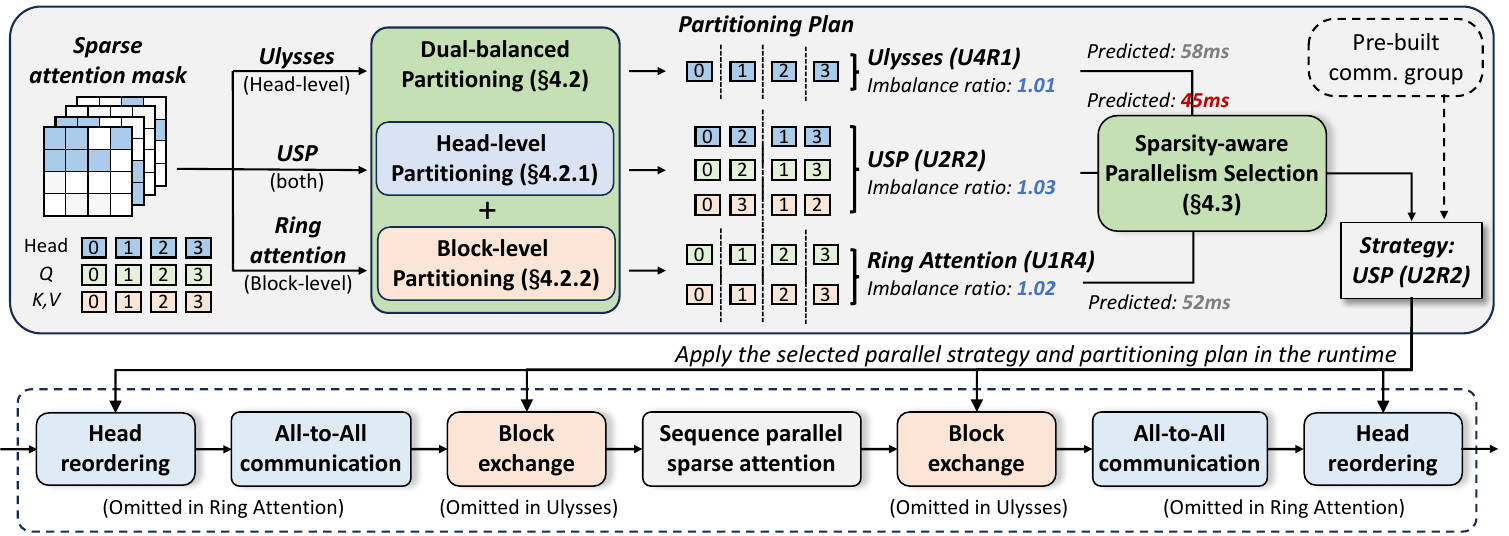}
    \vspace{-1.8em}
    \caption{The overview of \nickname. For every attention computation, each parallel strategy applies its specific partitioning method to mitigate the workload imbalance, achieving a post-balancing sparse imbalance ratio close to 1. Furthermore, \nickname leverages a sparsity-aware selection mechanism that considers the parallel strategy $\bf{p}$, sparse pattern $\bf{s}$, and \textit{sparse imbalance ratio} $\rho_s$ to predict and apply the optimal parallel strategy and the partitioning plan at runtime.}
    \vspace{-1.5em}
    \label{fig:overview}
\end{figure*}

\subsection{Challenges in Real Systems}
Although the workload imbalance can be theoretically addressed, there remain several challenges. 
\vspace{-0.8em}
\begin{itemize}
    \item \textbf{How to design a general dual-balanced parallelism method.} The sparse mask exhibits workload imbalance at both head and block levels. A key challenge is to decouple their interplay and design a general dual-balanced method that is agnostic to any block-wise sparse mask pattern.
    \vspace{-0.5em}
    \item \textbf{How to mitigate the overhead brought by workload balancing.} Workload balancing inevitably introduces overhead from extra reordering or data exchange. The challenge lies in mitigating the overhead to prevent it from negating the performance benefits.
    \vspace{-0.5em}
    \item \textbf{How to choose parallelism strategy against sparse dynamics.} As discussed in \S\ref{sec:background}, Ulysses and Ring Attention each has distinct advantages, and the choice of their degrees in constructing sequence parallelism impacts performance. The impact is more pronounced with sparse attention, where the sparse mask varies across denoising steps or even individual transformer layers, and efficiently determining the parallelism strategy presents a notable challenge.
\end{itemize}

\section{Methodology}\label{sec:method}
% \hk{[~2.5 pages.]}
% As concluded in Sec. 3, it is important to introduce a technique that can balance the workload on Multi-GPU without heavy overhead. We propose a Dual-Balanced Sequence Parallelism(db-SP), a widely used method to balance the workload of sparse attention in any parallelism strategies.

To address the challenges in building such a dual-balanced system that incorporates both sequence parallelism and sparse attention, we propose \nickname. In this section, we first introduce the dual-balanced method in \nickname that efficiently balances the workload for Ulysses, Ring Attention, and USP with negligible overhead, and then propose a sparsity-aware parallelism selection mechanism to determine the parallelism strategy for each transformer layer. 

\subsection{Overview}
% \hk{Describe the system overview according to the Figure. }

The overview of \nickname is shown in the Fig.~\ref{fig:overview}. When computing attention across GPUs, sequence parallelism strategies including Ulysses, Ring Attention, or the hybrid USP (U$x$R$y$) are available, where U$x$R$y$ denotes that every $x$ GPUs form a Ulysses group (head-level parallelism), whereas every $y$ GPUs form a Ring Attention group (Block-level parallelism). Each strategy applies its specific partitioning to mitigate the dual-level (\textit{i.e.}, head and block) workload imbalance, achieving a post-balancing \textit{sparse imbalance ratio} $\rho_s$ close to 1. Taking the parallel strategy $\bf{p}$, the sparse attention pattern $\bf{s}$, and the resulting $\rho_s$ as inputs, \nickname employs a sparsity-aware parallelism selection mechanism to predict the optimal strategy and apply its corresponding partitioning scheme for runtime deployment.

% According to different parallelism strategies, we design a tailored partitioning method to alleviate the problem of the workload imbalance. Then, according to the partitioning plan, we use a formula to predict the runtime of different parallelism strategies and choose the best one. During the inference, we apply the partitioning plan to the queries, keys, values, and hidden states to achieve the correctness.

\subsection{Dual-Balanced Partitioning}
Given a random block-wise sparse mask, the generated dual-balanced partitioning plan results in a \textit{sparse imbalance ratio} $\rho_s$ close to 1. In this section, we first introduce the head-level partitioning for Ulysses and the block-level partitioning for Ring Attention, respectively. For each, we detail the partitioning approach and the corresponding overhead mitigation method. Based on that, we clarify how to incorporate both for USP. 

\begin{figure}[t]
\label{fig:headlevel}
\includegraphics[width=1\columnwidth]{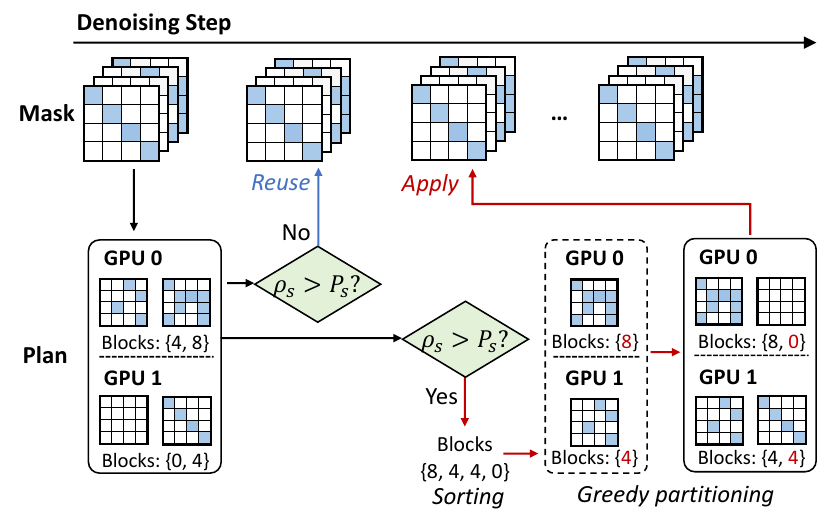}
\vspace{-2.0em}
\caption{Head-level partitioning in \nickname.}
\vspace{-1.5em}
\end{figure}

\subsubsection{Head-level Partitioning}
% (\hk{'level' maybe more concise.})
To achieve balance, head-level partitioning distributes the attention heads to different GPUs, so that on every GPU, the total number of dense blocks in the sparse masks corresponding to its attention heads is approximately the same. Since there is only one synchronization point across GPUs (\textit{i.e.}, the All-to-All after attention), the formulation of the \textit{sparse imbalance ratio} $\rho_s$ (Eq.~\ref{eq:ratio}) can be specified as follows.
\begin{align}
\rho_s = \frac{\max_{g\in\mathcal{G}}(b_{0, g}({\bf{s}}))}{\sum_{g\in\mathcal{G}} b_{0,g}({\bf{s}}) / |\mathcal{G}|}, b_{0,g}({\bf{s}}) = \sum_{j\in \mathcal{H}_g} b^{\text{head}}_j({\bf{s}}),
\end{align}
where $\mathcal{H}_g$ denotes the partitioned head set on GPU $g$, and $b^{\text{head}}_j({\bf{s}})$ is the block number of $j$-th head's sparse mask. 
% Head-level partitioning mitigates head-level workload imbalance in sparse attention by partitioning heads using a greedy algorithm and, for dynamic sparsity, reusing partitioning plans across timesteps when the imbalance ratio exceeds a threshold.

\textbf{Partitioning Approach.}
We apply a straightforward greedy algorithm for head-level partitioning. Specifically, we first sort the heads in descending order based on the number of dense blocks. Then, we iteratively assign each sorted head to a GPU according to the following principle: each head is always placed onto the GPU that currently has the least cumulative number of blocks. Such a loop continues until all heads are assigned. The sufficiently large number of heads and number of blocks per head in visual generative models make the simple partitioning method highly effective, resulting in a $\rho_s \leq$ 1.1 with Ulysses, which is detailed in \S\ref{sec:ablation}. 
% Ulysses splits the attention computation at the head level. As concluded in Sec. 3, the sparse pattern varies a lot between heads, so the workload is inevitably imbalanced. Intuitively, altering the order of heads can alleviate the head-level imbalance. Based on this idea, we propose a simple algorithm to determine the partitioning plan. The algorithm uses the ideology of the greedy algorithm. First, we accumulate the blocks that need to be calculated on each head, which can be regarded as the workload of each head. Then we sort the sums in descending order and rearrange them into different groups in turn. The process of sorting obeys the following rules: choose the group that owns the smallest head sums and has not been filled yet. We record the rearranged order and pass it as a parameter into the model's inference process. In this way, the workload on each GPU will be approximately. 

\textbf{Overhead Mitigation.}
Attention masks in each head pose similarity across denoising steps, thereby creating considerable potential for minimizing the overhead of the head-level partitioning. For a dynamic sparse attention method such as SpargeAttn~\cite{zhang2025spargeattn}, the sparse masks are generated online, and thus the partitioning plan is determined at runtime for each attention computation, with the number of partitioning executions equaling the number of layers multiplied by the number of denoising steps (\textit{e.g.}, 2,000 times in Wan2.1-T2V-14B). To this end, we propose a \textbf{\textit{partitioning plan reusing mechanism by exploiting the similarity across denoising steps}}. 

% As shown in Fig.~\ref{}, the sparse pattern in the attention mask undergoes significant changes in the early steps but remains largely unchanged in the later steps. 
Based on the observation, we introduce a threshold $P_s$ to automatically determine the timing for generating a new partitioning plan. For a certain transformer layer, we reuse the partitioning plan from the previous denoising step if the current $\rho_s$ falls below the predefined threshold $P_s$, otherwise, a new plan is generated based on the current sparse mask. With $P_s$ set to 1.10, partitioning is only necessary in 5 out of 50 denoising steps with Wan2.1-T2V-14B.

\subsubsection{Block-level Partitioning}

% Token-level partitioning balances workload across tokens in multiple dimensions and introduces a parameter to minimize inter-device communication.

Block-level partitioning is responsible for distributing the $Q,K,V$ chunks onto different GPUs, so that on every GPU at every iteration, the computed number of dense blocks is approximately the same. As discussed in \S\ref{sec:motivation}, Ring Attention performs synchronization across GPUs via Ring P2P communication at every iteration, the formulation of the \textit{sparse imbalance ratio} $\rho_s$ (Eq.~\ref{eq:ratio}) is specified by
\begin{align}
\rho_s = \frac{\sum_{i\in \{0, ..,|\mathcal{G}|-1\}}(\max_{g\in\mathcal{G}}(b_{i, g}({\bf{s}})))}{\sum_{i\in \{0, ..,|\mathcal{G}|-1\},g\in\mathcal{G}} b_{i,g}({\bf{s}}) / |\mathcal{G}|}, 
\end{align}
where $b_{i,g}({\bf{s}})$ is the dense block number in the computation of the $g$-th $Q$ chunk and the $((i+g) \bmod |\mathcal{G}|)$-th $K,V$ chunk accumulated across all attention heads. 

\begin{figure}[t]
\label{fig:tokenlevel}
\includegraphics[width=1\columnwidth]{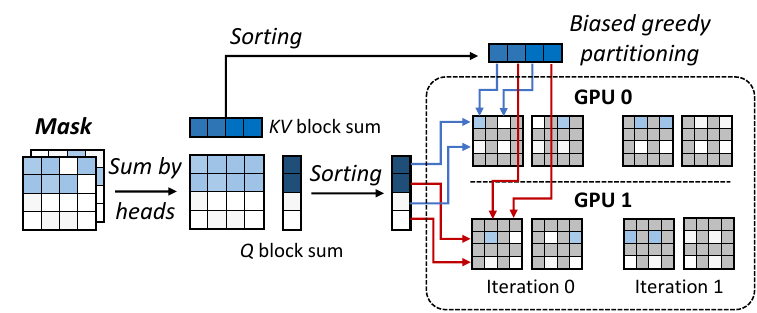}
\vspace{-1.8em}
\caption{Block-level partitioning in \nickname.}
\vspace{-1.5em}
\end{figure}

\textbf{Partitioning Approach.} 
We quantify the workload of Ring Attention by the number of dense blocks in the 2D computation between $Q$ and $K,V$. Each GPU is assigned a specific $Q$ chunk, while each iteration processes a specific $K,V$ chunk. Therefore, the goal is to partition the sparse mask as evenly as possible into $|\mathcal{G}| \times |\mathcal{G}|$ regions, each defined by a pair of $Q$ chunk and $K,V$ chunk. To achieve that, we also apply a greedy algorithm to partition $Q$ across GPUs and $K,V$ across iterations, respectively. Given a block-wise sparse mask, we partition $Q,K,V$ at the block-size granularity (\textit{e.g.}, 64 as in PAROAttention). For $Q$, we sort each block in descending order by the total count of its corresponding $K,V$ blocks and then greedily assign them across GPUs, putting every $Q$ block onto the GPU with the least accumulated block number. Similarly, for $K,V$, we sort each block by its number of corresponding $Q$ blocks and greedily assign them across iterations. Such an approach yields a $\rho_s \leq$1.05 with Ring Attention, as detailed in \S\ref{sec:overhead}. 

% Compared with Ulysses, ring attention splits the attention computation into a more elaborate granularity. Although it maintains the splitting strategy in the dimension of sequence, its ring communication introduces significant complexity by requiring multiple, synchronized iterations to complete a single attention operation. This fine-grained pipelining creates a tight dependency chain across all devices, where the slowest device in the ring dictates the overall speed, making the system highly sensitive to latency and imbalances.

% Inspired by the head-level partitioning algorithm, we propose a similar algorithm leveraging the ideology of the greedy algorithm. First, we accumulate the blocks of the attention map that need to be computed in the dimension of the head. As we have processed the imbalance in the dimension of head, this process will achieve a better performance globally. Then, we accumulate the computation workload in the dimensions of row and column, respectively. The third step is similar to the strategy used in head-level partitioning. We sort the sums in descending order and rearrange them into different groups in turn. The process of sorting obeys the following rules: choose the group that owns the smallest accumulated sum and has not been filled yet. 

\textbf{Overhead Mitigation.} 
Compared to head-level partitioning, block-level partitioning involves not only the intra-GPU data reorderings but also the inter-GPU data exchanges. Such data exchange is necessitated by the workload balancing. Although the $Q$ and $K,V$ for a given token are generated on a certain GPU, they are subsequently transferred to other GPUs for computation to achieve a balanced workload. Inter-GPU communication is much expensive than intra-GPU data reordering, and thus the key of overhead mitigation lies in minimizing the inter-GPU data exchanges. 

We introduce \textbf{\textit{a reward factor $R_b$ to encourage the $\bm{Q}$ and $\bm{K,V}$ to reside on the initial GPU}}. 
% Specifically, when performing the greedy algorithm to partition $Q$ blocks, the accumulated block number corresponding to the orginal GPU is biased by being substracted by $R_b$ times of the accumulated $K,V$ blocks corresponding to the current $Q$ block, and the sorting procedure is conducted to find the GPU with the least accumulated block number. 
Taking partitioning $Q$ blocks as an example, the modified greedy algorithm proceeds as follows for each $Q$ block. First, the accumulated $K,V$ block count on the $Q$ block's initial GPU is artificially reduced by the reward $R_b$ multiplied by the current $Q$ block's $K,V$ block count. Then, the algorithm selects the GPU with the least resulting accumulated count for assignment. Similarly, the $K,V$ blocks are partitioned in the biased greedy manner. 
% \textcolor{purple}{See \S~\ref{sec:overhead} for more details.}
% Such a reward 
% Moreover, to reduce the potential overhead of token exchange between different devices, we introduce a new parameter to encourage the tokens to stay in the original devices. In this way, the communication overhead can be reduced. We record the rearranged order and pass it as a parameter into the model's inference process. The framework of head-level partitioning is shown in the Figure\ref{fig:tokenlevel}. 

% \begin{algorithm}[tb]
%    \caption{Token-level partitioning}
%    \label{alg:example}
% \begin{algorithmic}
%    \STATE {\bfseries Input:} data $sparse\_mask$, data $balance\_reward$
%    \STATE Accumulate the sparse in the dimension of head
%    \STATE Accumulate the sparse in the dimension of row
%    \REPEAT 
%    \STATE Put the remaining maximum into the smallest group considering balance\_reward
%    \STATE Record the new row-level partitioning plan 
%    \UNTIL{Finish partitioning}
%    \STATE Accumulate the sparse in the dimension of column
%    \REPEAT 
%    \STATE Put the remaining maximum into the smallest group, considering balance\_reward
%    \STATE Record the new column-level partitioning plan 
%    \UNTIL{Finish partitioning}
%    \STATE Record the new partitioning plan and convert format
%    \STATE Implement the partitioning plan before computation
% \end{algorithmic}
% \end{algorithm}

\begin{algorithm}[t]
   \caption{Dual-Balancing Partitioning}
   \label{alg:dualbalance}
\begin{algorithmic}[1]
   \STATE {\bfseries Input:} sparse pattern $\bf{s}$, parallel strategy ${\bf{p}}$ (U$x$R$y$), reusing threshold $P_s$, partitioning reward $R_b$
   % $ulysses\_degree$, data $ring\_degree$, data $sparse\_mask$, data $reward$
    \STATE \textit{head\_partitioning\_plan} $\gets$ \textit{old\_head\_partitioning\_plan}
    \STATE \textcolor{blue}{// Head-level partitioning}
    \IF{($x>$  1) and (cal\_imbalance\_ratio(${\bf{s}}$) $> P_s$) }
    % \STATE \textcolor{blue}{// Check if reuse the partitioning plan}
    % \IF{cal\_imbalance\_ratio(${\bf{s}}$) $> P_s$ }
    \STATE \textit{blocks\_per\_head} $\gets$ get\_blocks\_per\_head($\bf{s}$)
    % \STATE $head\_groups = [[]\ for\ \_\ in\ range(ulysses\_degree)]$
    % \STATE $head\_order = argsort(sparse\_mask.sum((-1,-2)))$
    \STATE \textit{sorted\_head\_id} $\gets$ desc\_argsort(\textit{blocks\_per\_head})
    % \FOR{$idx\ in\ range(head\_order.shape)$}
    \STATE \textit{sum\_per\_gpu} $\gets$ $\{0\}_x$
    \FOR{\textit{head\_id} in enumerate(\textit{sorted\_head\_id})}
    \STATE \textit{gpu\_id} $\gets$ argmin(\textit{sum\_per\_gpu})
    \STATE \textit{head\_partitioning\_plan}
    [\textit{gpu\_id}].append(\textit{head\_id})
    \STATE \textit{sum\_per\_gpu}.update()
    \ENDFOR    
    % \ENDIF
    % \STATE $head\_level\_partitioning\_plan =$
    % \STATE \quad $[i\ for\ g\ in\ head\_groups\ for\ i\ in\ g]$
    \ENDIF

    \STATE \textit{q\_partitioning\_plan},\textit{kv\_partitioning\_plan} $\gets \emptyset$
    % \STATE \textit{kv\_partitioning\_plan} $\gets \emptyset$
    \STATE \textcolor{blue}{// Block-level partitioning}
    \IF{$y>$ 1 }
    % \STATE $head\_sum=sparse\_mask.sum(0)$
    \STATE \textcolor{blue}{// Assume workload is balanced across heads}
    \STATE \textit{blocks\_accum\_by\_head} $\gets$ sum\_by\_head(${\bf{s}}$)
    % \STATE \textcolor{blue}{// For each $Q$ block, sum up $K,V$ blocks}
    % \STATE $row\_sum = head\_sum.sum(1)$
    \STATE \textit{blocks\_per\_q} $\gets$ sum\_by\_kv(\textit{blocks\_accum\_by\_head})
    % \STATE \textit{blocks\_per\_kv} $\gets$ sum\_by\_q(\textit{blocks\_accum\_by\_head})
    % \STATE $col\_sum = head\_sum.sum(0)$
    % \STATE $row\_groups = [[]\ for\ \_\ in\ range(ring\_degree)]$
    % \STATE $row\_order = argsort(row\_sum)$
    \STATE \textit{sorted\_q\_id} $\gets$ desc\_argsort(\textit{blocks\_per\_q})
    % \STATE \textit{sorted\_kv\_id} $\gets$ desc\_argsort(\textit{blocks\_per\_kv})
    % \FOR{$idx\ in\ range(row\_order.shape)$}
    \STATE \textit{sum\_per\_gpu} $\gets$ $\{0\}_y$
    \FOR{\textit{q\_id} in enumerate(\textit{sorted\_q\_id})}
    % \STATE $gid=argmin(group\_sums-$
    % \STATE \quad $reward\cdot\delta(origin\_group-new\_group))$
    \STATE \textcolor{blue}{// Perform biased greedy partitioning with reward}
    \STATE \textit{biased\_sum\_per\_gpu} $\gets$ [$s- R_b$ if $g$ is        initial\_gpu
    \STATE \quad for $g,s$ in enumerate(\textit{sum\_per\_gpu})]
    \STATE  \textit{gpu\_id} $\gets$ argmin(\textit{biased\_sum\_per\_gpu})
    % \STATE$row\_groups[gid].append(idx)$
    \STATE \textit{q\_partitioning\_plan}[\textit{gpu\_id}].append(\textit{q\_id})
    \STATE \textit{sum\_per\_gpu}.update()
    \ENDFOR
    % \STATE $row\_level\_partitioning\_plan =$
    % \STATE \quad $[i\ for\ g\ in\ row\_groups\ for\ i\ in\ g]$
    \STATE \textcolor{blue}{// Repeat to partition $K,V$ blocks}
    \STATE \textit{blocks\_per\_kv} $\gets$ sum\_by\_q(\textit{blocks\_accum\_by\_head})
    \STATE ...
    % \STATE $col\_sum = head\_sum.sum(0)$
    % \STATE $col\_groups = [[]\ for\ \_\ in\ range(ring\_degree)]$
    % \STATE $col\_order = argsort(col\_sum)$
    % \FOR{$idx\ in\ range(col\_order.shape)$}
    % \STATE $gid=argmin(group\_sums-$
    % \STATE \quad $reward\cdot\delta(origin\_group-new\_group))$
    % \STATE $col\_groups[gid].append(idx)$
    % \ENDFOR
    % \STATE $col\_level\_partitioning\_plan =$
    % \STATE \quad $[i\ for\ g\ in\ col\_groups\ for\ i\ in\ g]$
    % \STATE $balanced\_sparse\_mask=$
    % \STATE \quad $sparse.index\_select(partitioning\_plans)$
    \ENDIF
    \STATE {\bfseries Output:} \textit{head\_partitioning\_plan}, \textit{q\_partitioning\_plan}, \textit{kv\_partitioning\_plan}
    
\end{algorithmic}
\end{algorithm}

\subsubsection{Dual-Balancing Partitioning Algorithm}

The optimization of head-level partitioning and block-level partitioning exhibits interdependence. Specifically, the partitioning plan of one determines the total number of blocks and the sparse pattern for the other. To avoid the prohibitive search cost associated with joint optimization, we simplify the problem based on a key insight: both the proposed head-level and block-level partitioning approaches individually yield a near-perfectly balanced workload across GPUs. Consequently, \textbf{\textit{the dual-level optimization problem can be decomposed into two subproblems}}, each targeting the imbalance minimization at its respective level based on the assumption that the other is well-balanced.

In practice, we choose to perform head-level partitioning first. Under the assumption that the workload is perfectly balanced across all GPUs within the same Ulysses group, an identical block-level partitioning plan can be optimally derived for all Ring Attention groups. The detailed dual-balancing partitioning algorithm is presented in Alg.~\ref{alg:dualbalance}. At the head level, if the partitioning plan from the last denoising step fails to be reused, the algorithm performs greedy head-level partitioning to generate a new one (Lines 4-13). At the block level, we assume workload is perfectly balanced across different heads, and eliminate the influence of the head dimension through summation. During the partitioning procedure, the reward $R_b$ is introduced to address the initial GPU residence preference of the $Q$ and $K,V$ (Lines 23-25), in order to mitigate the inter-GPU data exchange overhead. 

The general principles of hyperparameter selection are as follows. First, on platforms with lower interconnect bandwidth, it is beneficial to use a larger $R_b$ to reduce inter-GPU data exchange overhead. Moreover, with higher sparsity, the imbalance across GPUs is more pronounced. Accordingly, $R_b$ and $P_s$ can be adjusted to smaller values for better workload balancing. In our experiments, the hyperparameters are empirically selected with $R_b$=5 and $P_s$=1.05. Based on the sensitivity analysis presented in \S \ref{sec:overhead}, performance remains largely unaffected across ranges of $R_b$=5$\sim$10 and $P_s$=1.03$\sim$1.10.

\subsection{Sparsity-Aware Parallel Strategy Selection}
As discussed in \S\ref{sec:background}, the employment of USP~\cite{fang2024uspunifiedsequenceparallelism} forms a parallel strategy $\bf{p}$ (U$x$R$y$) design space to incorporate Ulysses and Ring Attention. Given hardware, model settings, and inputs, a fixed parallel strategy can be optimal for dense attention under static conditions. Nevertheless, the sparse attention exhibits dynamic sparsity patterns $\bf{s}$ that vary per transformer layer and denoising step, which influences both the \textit{sparse imbalance ratio} $\rho_s$ and the efficiency of a parallel strategy $\bf{p}$ itself, leading to performance degradation with a fixed parallel strategy. 

To further investigate the influence of the sparse pattern $\bf{s}$, given parallel strategy $\bf{p}$ (U$x$R$y$), we formulate the latency of the multi-GPU attention part $\mathcal{L}(\bf{s}, \bf{p})$ by
\vspace{-0.5em}
\begin{equation}\label{eq:lat}
\begin{split}
\mathcal{L}(\bf{s}, \bf{p}) = & \mathcal{L}^{\text{all2all}}(x({\bf{p}})) + \mathcal{L^{\text{attn}}}(\bf{s}, \bf{p}), \\
\mathcal{L^{\text{attn}}}(\bf{s}, \bf{p}) = &
\left[ \max\bigl(\mathcal{L}^{\text{comp}}({\bf{s}}), \mathcal{L}^{\text{p2p}}(y({\bf{p}})) \bigr) \right. \\
& \quad \left. \times (y({\bf{p}})-1) + \mathcal{L^{\text{comp}}}({\bf{s}}) \right] \times \rho_s ({\bf{s, p}}), \\
\mathcal{L}^{\text{comp}}({\bf{s}}) = & (\frac{\mathcal{L}^{\text{dense}}}{|\mathcal{G}|} \times \text{density(\bf{s})})+\mathcal{L}^{\text{launch}}.
\end{split}
\end{equation}
\vspace{-1.5em}

In Eq.~\ref{eq:lat}, $\mathcal{L}^{\text{all2all}}(x({\bf{p}}))$ denotes the All-to-All communication latency, and hence is determined by the degree of Ulysses. Note that the sparse pattern $\bf{s}$ does not affect the communication volume. $\mathcal{L^{\text{attn}}}(\bf{s}, \bf{p})$ represents the attention latency, which considers the computation-communication overlap in Ring Attention, where $\mathcal{L}^{\text{comp}}({\bf{s}})$ is the attention computation latency and $\mathcal{L}^{\text{p2p}}(y({\bf{p}}))$ is the Ring P2P communication latency at one iteration. $\mathcal{L^{\text{attn}}}(\bf{s}, \bf{p})$ also includes the performance degradation brought by workload imbalance via the multiplicative factor $\rho_s ({\bf{s, p}})$. Moreover, the computation latency $\mathcal{L}^{\text{comp}}({\bf{s}})$ is determined by the sparsity/density, and captures the kernel launch overhead $\mathcal{L}^{\text{launch}}$. 

\begin{figure*}[t]
    \centering
    \includegraphics[width=0.95\textwidth]{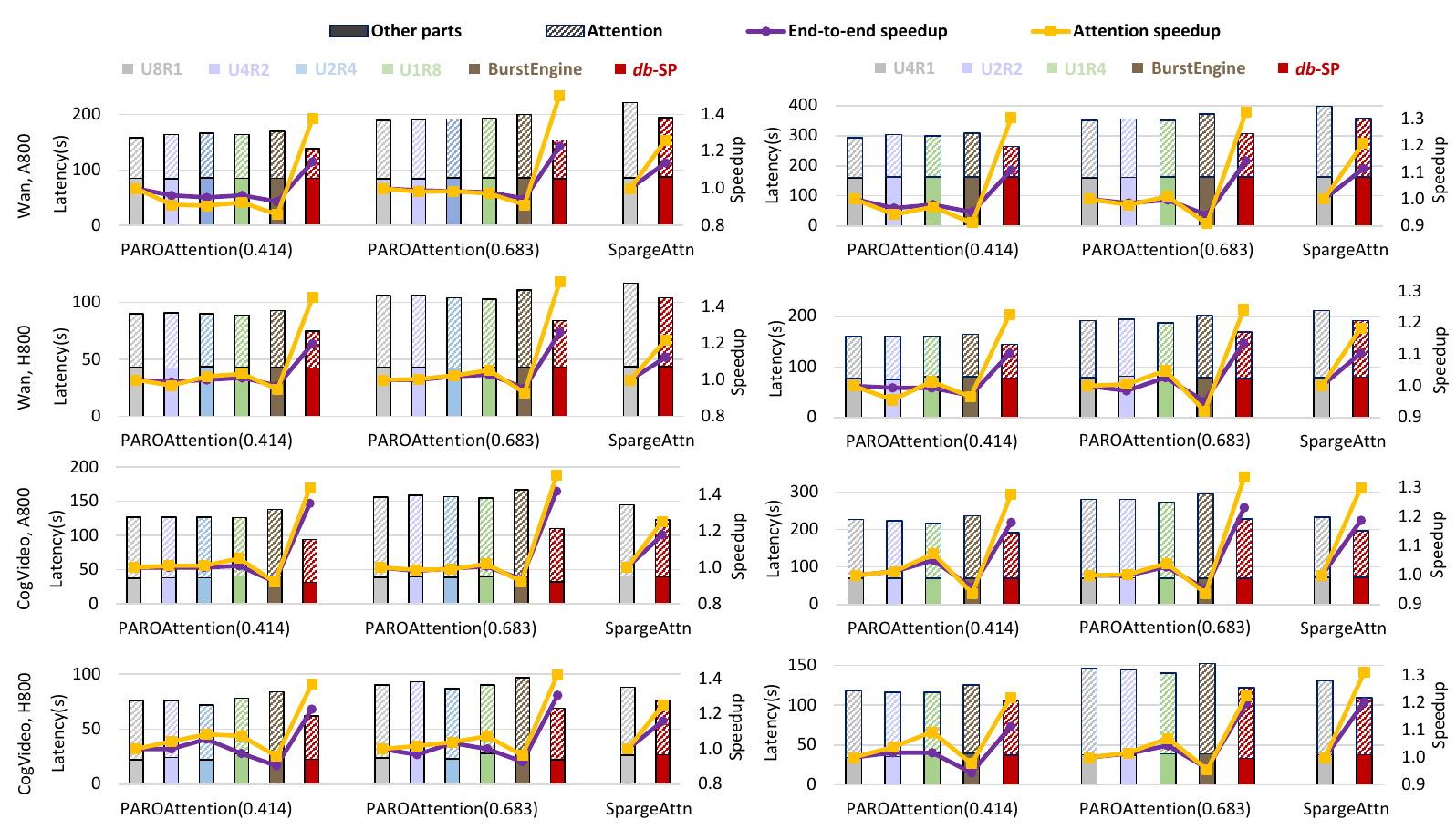}
    \vspace{-1.2em}
    \caption{The end-to-end latency comparison on eight (\textit{left}) and four (\textit{right}) GPUs. SpargeAttn only supports Ulysses due to the attention kernel implementation.}
    \vspace{-1.2em}
    \label{fig:results}
\end{figure*}

On top of that, \nickname utilizes a layer-wise sparsity-aware parallel strategy selection mechanism, enabling the strategy switch for every attention computation. The core insight is that, sequence parallelism avoids partitioning model weights across GPUs, \textbf{\textit{switching between different sequence parallel strategies requires no model reloading, enabling dynamic strategy selection at runtime}}.  The switching necessitates changes in the inter-GPU communication patterns, which require the reconfiguration of communication groups. To eliminate the overhead, \nickname introduces pre-built communication groups, constructing all potential communication groups ($\log_2(|\mathcal{G}|)+1$ in total) for different parallel strategies before runtime. 

Given a sparse pattern $\bf{s}$, we predict the latency of each parallel strategy $\bf{p}$ using Eq.~\ref{eq:lat}. Specifically, $\mathcal{L}^{\text{all2all}}(x)$, $\mathcal{L}^{\text{p2p}}(y)$, $\mathcal{L}^{\text{dense}}$, and $\mathcal{L}^{\text{launch}}$ are fitted offline using profiling data collected from dense attention computations. Notably, both $\mathcal{L}^{\text{all2all}}(x)$ and $\mathcal{L}^{\text{p2p}}(y)$ are platform-specific and must be individually fitted for each distinct hardware configuration, including its network topology. These two functions are independent of sparsity, which allows them to be fitted offline. In the event of network fabric congestion, the two functions require refitting, and to mitigate prediction errors, periodic fitting can be introduced to track real‑time network conditions. At runtime, the parallel strategy with the lowest predicted latency is selected and executed using its pre‑built communication groups ($x$ GPUs for Ulysses and $y$ GPUs for Ring Attention), as illustrated in Fig.~\ref{fig:overview}. 
\section{Implementation}
% We prototype db-SP as an end-to-end framework and an independent operator testbench. To demonstrate the wide validity of our method, we customize the end-to-end framework based on both ParaAttention and xDiT, which supports different sequence parallelism strategies. As for models, we choose Wan2.1 and CogVideoX1.5. Meanwhile, the operator testbench is also supported, which is customized based on USP. 

 We implement \nickname based on the open-source code of USP~\cite{fang2024uspunifiedsequenceparallelism}, and replace the attention kernel backend with the implementation of PAROAttention~\cite{zhao2025paroattentionpatternawarereorderingefficient} or SpargeAttn~\cite{zhang2025spargeattn} depending on the sparse method, to support block-wise sparse masks as input. \nickname contains interfaces to integrate into DiT inference engines for sequence parallel attention computation. Currently, we have integrated \nickname into the mainstream inference engines, including xDiT~\cite{fang2024xdit} and ParaAttention~\cite{wavespeed2025paraattention}, to support the end-to-end inference of different models. \nickname involves intra-GPU data reorderings and inter-GPU data exchanges in the partitioning approach, which are implemented by \texttt{index\_select} and \texttt{all\_to\_all} with NCCL~\cite{NCCL} backend in PyTorch~\cite{paszke2019pytorch}, respectively.

\section{Evaluation}
% \hk{[~2.5 pages.]}
We evaluate \nickname with various sparse masks, models, and hardware platforms. The evaluation demonstrates that \nickname yields an attention speedup of 1.40× and an end-to-end speedup of 1.25$\times$ on average. Meanwhile, \nickname introduces merely less than 5\% partitioning overhead.

% \textbf{Testbed.} We experiment on two distinct GPU server configurations: NVIDIA H800 and NVIDIA A800 platforms. \emph{The H800 platform} features 8 GPUs with a homogeneous fully-connected NVLink fabric (NV8). All GPUs exhibit identical connectivity patterns, with each GPU having direct PIX (single PCIe bridge) access to its corresponding Mellanox ConnectX-6 NIC and SYS connectivity to others. \emph{A800 Platform} configuration presents a more complex hierarchical topology. While maintaining the fully-connected NVLink (NV8) among all 8 GPUs, the NIC connectivity exhibits NUMA-aware segmentation.
\subsection{Setup}
\textbf{Testbed.} We conduct experiments on two distinct GPU servers. The first is equipped with eight NVIDIA A800 80GB SXM4 GPUs, and the second has eight NVIDIA H800 80GB GPUs. The GPUs in both servers are fully connected via the NVLink high-speed interconnect fabric. The corresponding software environment includes CUDA 12.1~\cite{nvidia2024cuda}, PyTorch 2.5.1~\cite{paszke2019pytorch}, and NCCL 2.21.5~\cite{NCCL}. 

\textbf{Benchmark.} We evaluate \nickname with two models, Wan2.1-T2V-14B~\cite{wan2025wanopenadvancedlargescale} and CogVideoX1.5-5B~\cite{yang2025cogvideoxtexttovideodiffusionmodels}, abbreviated as Wan2.1 and CogVideoX1.5 in the following. We evaluate by generating 81-frame 1280P videos with 50 sampling steps, and 161-frame 1280P videos with 50 sampling steps, for the Wan2.1-T2V-14B model and the CogVideoX1.5-5B model, respectively. 
% After being tokenized by 3D-VAE, Wan2.1 generates 21 frames with 3600 tokens per frame, while CogVideoX processes 21 frames with 4080 tokens per frame.
The 3D-VAE tokenizer processes the input into a sequence of tokens, upon which Wan2.1-T2V-14B generates 21 frames at 3600 tokens per frame, and CogVideoX1.5-5B produces 21 frames at 4080 tokens per frame.

For the sparse attention method, we use PAROAttention~\cite{zhao2025paroattentionpatternawarereorderingefficient} to represent the ones with static masks across denoising steps, and use SpargeAttn~\cite{zhang2025spargeattn} for the dynamic ones. Specifically, for PAROAttention~\cite{zhao2025paroattentionpatternawarereorderingefficient}, we use two sparse masks with sparsity of 0.414 and 0.683, denoted as PAROAttention(0.414) and PAROAttention(0.683) in the following, respectively. Note that the attention kernel of SpargeAttn only supports Ulysses. 

%: GPUs 0-1 connect via PXB to NICs 0-1 and via NODE to NICs 2-3; GPUs 2-3 connect via PXB to NICs 2-3 and via NODE to NICs 0-1; GPUs 4-5 connect via PXB to NICs 5-6; GPUs 6-7 connect via PXB to NICs 7-8. 

% \textbf{Benchmarks.} For sparse attention, we select the most typical ones from static sparse attention mechanisms and dynamic sparse attention mechanisms: PAROAttention and SpargeAttn. 

% \textbf{Baselines.} For sequence parallelism strategies, we select ring attention(U1R8), Ulysses(U8R1), and different settings of USP(U2R4 and U4R2). We also implement BurstEngine for baseline.
\textbf{Baseline.} We compare \nickname with the typical sequence parallelism methods including Ulysses~\cite{jacobs2023deepspeedulyssesoptimizationsenabling}, Ring Attention~\cite{liu2023ringattentionblockwisetransformers}, and USP~\cite{fang2024uspunifiedsequenceparallelism}. The parallel strategy employed in USP is specified as U$x$R$y$, indicating that the parallel degrees for Ulysses and Ring Attention are $x$ and $y$, respectively. Additionally, we employ BurstEngine~\cite{sun2025burstengineefficientdistributedframework} as our sparsity-aware sequence parallelism baseline, as it specifically optimizes workload imbalance for models utilizing block-wise sparse attention with Ring Attention.

\textbf{Evaluation Metrics.} We record the end-to-end inference latency and the attention latency for comparison. Also, we use the \textit{sparsity imbalance ratio} to measure the extent of workload balance across GPUs. 

\subsection{End-to-end Performance}\label{sec:e2e}
\textbf{Single-Node Results.} Fig.~\ref{fig:results} shows the end-to-end latency and attention latency comparison on four and eight GPUs. 
\nickname consistently outperforms all baselines, achieving an average end-to-end speedup of 1.12-1.26$\times$ for Wan2.1 and 1.16-1.42$\times$ for CogVideoX1.5 on eight GPUs. On four GPUs, the end-to-end speedup ranges from 1.10$\times$ to 1.15$\times$ and from 1.10$\times$ to 1.15$\times$ for Wan2.1 and CogVideoX1.5, respectively. \nickname achieves higher speedup on eight GPUs, as the workload imbalance exacerbates with the increase in GPU count. 
The sparse attention computation still occupies most of the end-to-end latency, and \nickname delivers an average of 1.26$\times$/1.38$\times$ attention speedup over Ulysses and 1.22$\times$/1.42$\times$ over Ring Attention on four/eight GPUs. 
% We also measure the attention time in the end-to-end inference separately. The results show that the attention speedup is up to 1.5-1.6× and contributes to the end-to-end speedup. 
Notably, BurstEngine fails to outperform other baselines due to the performance degradation of the attention kernel. Taking a block size of 256 as an example, BurstEngine uses a tile size of 32 for the attention kernel on eight GPUs, which is 1.25$\times$ slower than using a tile size of 64. 
Moreover, we observe that the baselines with different parallel strategies perform similarly without workload balancing. However, once the workload imbalance is addressed, the performance gap becomes pronounced, which is detailed in \S\ref{sec:ablation}.

\textbf{Multi-Node Results.} To demonstrate the scalability of \nickname, we conduct experiments on two nodes, each equipped with eight A100 GPUs and connected via a 200 Gbps cross-node interconnect, using the PAROAttention sparse pattern. As we scale from single-node to multi-node deployment, the relative latency overhead of non-optimized communication becomes increasingly dominant. Nevertheless, the opportunity for improvement remains, as the workload imbalance across GPUs grows with the number of devices. As shown in Tab.~\ref{tab:multi}, \nickname achieves a 1.10× end-to-end speedup even in the multi-node setting.

\begin{table}[t]
\caption{The latency comparison on multi-node settings. U16R1 is not feasible due to the head number.}
\label{tab:multi}
\begin{center}
\setlength{\tabcolsep}{2pt} 
\resizebox{0.48\textwidth}{!}{
\begin{tabular}{lcccccc}
\toprule
 & U16R1 & U8R2 & U4R4 & U2R8 & U1R16 & \textit{db}-SP \\
\midrule
End-to-end & -- & 237s & 128s & 178s & Too slow & 116s (\textbf{1.10$\times$}) \\
\midrule
Attention & -- & 192s & 84.6s & 136s & Too slow & 73.6s (\textbf{1.15$\times$}) \\
\bottomrule
\end{tabular}
}
\end{center}
\vspace{-1.0em}
\end{table}

\begin{table}[t]
\centering
\small
\caption{Performance comparison between H100 and H800 GPUs. Results are collected with Wan2.1 and PAROAttention (0.683).}
\label{tab:gputypes}
\vspace{0.3em}
\setlength{\tabcolsep}{3pt}
\begin{tabular}{lcccccc}
\toprule
\multicolumn{7}{c}{8 GPUs} \\
\cmidrule(lr){1-7}
 & U8R1 & U4R2 & U2R4 & U1R8 & \nickname & Speedup \\
\midrule
H800 & 106s & 106s & 104s & 103s & 84.0s & 1.26$\times$ \\
H100 & 60.7s & 60.6s & 61.1s & 60.7s & 40.2s & \textbf{1.51$\times$} \\
\midrule
\multicolumn{7}{c}{4 GPUs} \\
\cmidrule(lr){1-7}
 & U4R1 & U2R2 & U1R4 & - & \nickname & Speedup \\
\midrule
H800 & 192s & 195s & 187s & - & 169s & 1.14$\times$ \\
H100 & 108s & 108s & 110s & - & 86.3s & \textbf{1.26$\times$} \\
\bottomrule
\end{tabular}
\end{table}

\begin{table*}[t]
\caption{Benefits breakdown with dynamic sparse attention, measured with SpargeAttn and Wan2.1.}
% \vspace{-1.0em}
\label{tab:spargeattn_results}
\begin{center}
\begin{small}
\vspace{0.5em}
\setlength{\tabcolsep}{2pt}
\begin{tabular}{lcccccc}
\toprule
 & U8R1 & U4R2 & U2R4 & U1R8 & \multicolumn{2}{c}{\textit{db}-SP} \\
\cmidrule(lr){6-7}
 & & & & & w/o dynamic parallel strategy (U4R2) & overall \\
\midrule
End-to-end & 230s & 238s & 282s & 306s & 214s (1.11$\times$) & 203s (1.13$\sim$1.51$\times$) \\
\midrule
Attention & 145s & 144s & 165s & 205s & 124s (1.16$\times$) & 117s (1.24$\sim$1.75$\times$) \\
\bottomrule
\end{tabular}
\end{small}
\end{center}
\vspace{-1.0em}
\end{table*}

% \begin{table}[t]
% \centering
% \caption{\textcolor{blue}{Performance comparison between H100 and H800 GPUs. Results collected with Wan2.1 and PAROAttention (0.683).}}
% \label{tab:gputypes}
% \vspace{0.5em}
% \begin{tabular}{l@{\hspace{4pt}}c@{\hspace{4pt}}c@{\hspace{4pt}}c@{\hspace{4pt}}c@{\hspace{4pt}}c@{\hspace{4pt}}c}
% \toprule
% \multicolumn{7}{c}{8GPUs} \\
% \cmidrule(lr){1-7}
% & U8R1 & U4R2 & U2R4 & U1R8 & \textit{db}-SP & speedup \\
% \midrule
% H800 & 90s & 91s & 90s & 89s & 75s & 1.2 \\
% H100 & 61s & 61s & 61s & 61s & 40s & 1.53 \\
% \midrule
% \multicolumn{7}{c}{4GPUs} \\
% \cmidrule(lr){1-7}
% & U4R1 & U2R2 & U1R4 & \textit{db}-SP & speedup \\
% \midrule
% H800 & 160s & 161s & 161s & 145s & 1.10 \\
% H100 & 108s & 108s & 110s & 86s & 1.26 \\
% \bottomrule
% \end{tabular}
% \end{table}

\textbf{Impact of GPU types.} Given the compute-intensive nature of long-sequence DiT workloads, the end-to-end performance gains delivered by \nickname are governed by the interplay between the GPU's interconnect bandwidth and its peak computational throughput (\textit{i.e.}, TFLOPS). 
\nickname accelerates the attention computation by dynamically balancing the workload across GPUs. Therefore, the achievable speedup is positively correlated with the proportion of attention computation time. As evidenced in Fig.~\ref{fig:results}, A800 GPUs yield a higher speedup than H800 GPUs, which can be attributed to A800 GPUs' lower computational throughput, which increases the relative proportion of time spent on computation, while the two platforms exhibit comparable interconnect performance.

A similar trend can be observed when comparing H100 and H800 GPUs. As presented in Tab.\ref{tab:gputypes}, \nickname attains a higher speedup on H100 GPUs, as the higher interconnect bandwidth on the H100 server reduces the portion of the unoptimized inter-GPU communication latency.

\subsection{Ablation Study}\label{sec:ablation}
% We ablate the head-level partitioning and the token-level partitioning, respectively. As shown in Table \ref{table:ablation}, applying either of the two partitioning plans can achieve acceleration, but the dual balance partitioning plan achieves the best theoretical optimization and actual acceleration. 

\begin{figure}[t]
    \centering
    \includegraphics[width=0.5\textwidth]{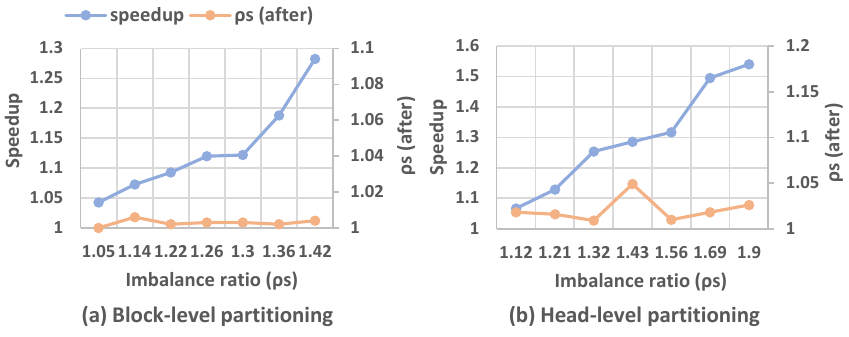}
    \vspace{-1.0em} % 也相应调整垂直间距
    \caption{Attention speedup as sparse imbalance ratio grows.}
    \vspace{0em}
    \label{fig:attentionspeedup}
\end{figure}

\textbf{Dual-balanced Partitioning.}
% To justify the potential speedup caused by the workload imbalance, we select several real sparse masks with different imbalance ratios from the different attention layers of Wan2.1. We measure the original time of attention and the time of applying head-level partitioning or token-level partitioning. As shown in Figure\ref{fig:attentionspeedup}, the speedup and imbalance ratio approximately showcase a direct proportion relationship, which highlights the efficiency of our approach over different cases.
We evaluate the attention speedup brought by head-level partitioning and block-level partitioning separately. Specifically, we use typical sparse masks with varying sparse imbalance ratios ranging across layers and steps and measure the corresponding attention speedups. As depicted in Fig.~\ref{fig:attentionspeedup}, the speedup achieved by the partitioning approach increases with the sparse imbalance ratio of the input mask, exhibiting a close-to-linear growth trend. Tab.~\ref{tab:dual_level_partitioning} shows the corresponding sparse imbalance ratios ($\rho_s$) after applying the proposed dual-level balancing approach. The achieved $\rho_s$ is within $6\%$ to the perfectly-balanced upper bound (\textit{i.e}., an imbalance ratio of 1).

\begin{table}[t]
\caption{The sparse imbalance ratio $\rho_s$ before and after dual-level partitioning, ranging across layers and steps.}
\label{tab:dual_level_partitioning}
\vspace{0.3em}
% \begin{small}
\resizebox{0.48\textwidth}{!}{
    \begin{tabular}{lcccc}
    \toprule
     & U8R1 & U4R2 & U2R4 & U1R8 \\
    \midrule
    $\rho_s$ before & 1.10$\sim$1.69 & 1.08$\sim$1.63 & 1.08$\sim$1.42 & 1.05$\sim$1.39 \\
    \midrule
    $\rho_s$ after & 1.00$\sim$1.06 & 1.00$\sim$1.01 & 1.00$\sim$1.01 & 1.00$\sim$1.01 \\
    \bottomrule
    \end{tabular}
}
% \end{small}
\end{table}

% \begin{table}[t]
% \caption{Ablation of head-level and token-level partitioning.}
% \label{table:ablation}
% \vskip 0.15in
% \begin{center}
% \begin{small}
% % \begin{sc}
% \begin{tabular}{lcccr}
% \toprule
% Parallelism strategy & Imbalance ratio & Speedup \\
% \midrule
% U2R4    & 1.95 & 1.00× \\
% \text{DB-SP (w/o token)}  & 1.29 & 1.98×\\
% \text{DB-SP (w/o head)}  & 1.44 & 1.5×\\
% \text{DB-SP (with both)}  & 1.00 & 2.48×\\
% \cmidrule{1-3}
% U4R2    & 2.24 & 1.00× \\
% \text{DB-SP (w/o token)}  & 1.28 & 1.49×\\
% \text{DB-SP (w/o head)}  & 1.61 & 1.11×\\
% \text{DB-SP (with both)}  & 1.00 & 2.81×\\

% \bottomrule
% \end{tabular}
% % \end{sc}
% \end{small}
% \end{center}
% \vspace{-1.0em}
% \end{table}

\begin{figure}[t]
    \centering
    \includegraphics[width=0.35\textwidth]{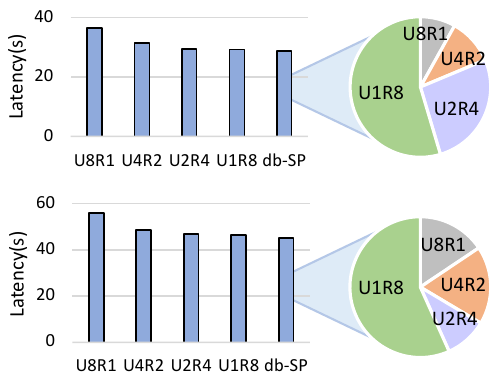}
    \vspace{-1.0em} % 也相应调整垂直间距
    \caption{Latency comparison to static parallel strategies. }
    \vspace{-1.0em}
    \label{fig:selection}
\end{figure}

\textbf{Benefits breakdown.} To further understand the benefits of \nickname, we modify the attention kernel of SpargeAttn \cite{zhang2025spargeattn} to support dynamic sparisty. Specifically, we adapt the kernel to Ring Attention computation by enabling \texttt{log\_sum\_exp} output. Experiments are conducted on the server with eight A100 GPUs. As shown in Tab.~\ref{tab:spargeattn_results}, the results demonstrate the effectiveness of both dual-level balancing and dynamic parallel strategy selection for dynamic sparse workloads.

\textbf{Sparsity-Aware Parallel Strategy Selection.}
We evaluate the benefits of the parallel strategy selection mechanism by comparing \nickname with different parallel strategies with workload balancing. The dynamic parallel strategy switching brings 1.02-1.27$\times$ end-to-end speedup compared to all the static strategies. To visualize the selection distribution, we present the detailed ratios of all the strategies in Fig.~\ref{fig:selection}.

\subsection{Overhead analysis}\label{sec:overhead}
\begin{figure}[t]
    \centering
    \includegraphics[width=0.5\textwidth]{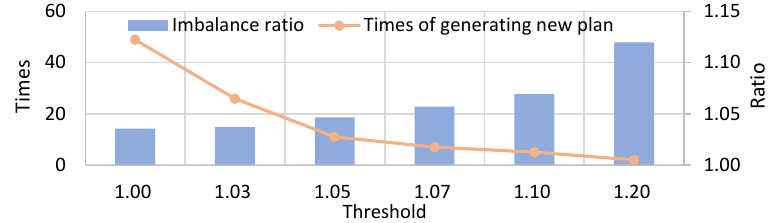}
    \vspace{-2em} % 也相应调整垂直间距
    \caption{Effect of threshold $P_s$ choice. }
    \vspace{-0.5em} % 也相应调整垂直间距
    \label{fig:thr}
\end{figure}

\begin{figure}[t]
    \includegraphics[width=0.5\textwidth]{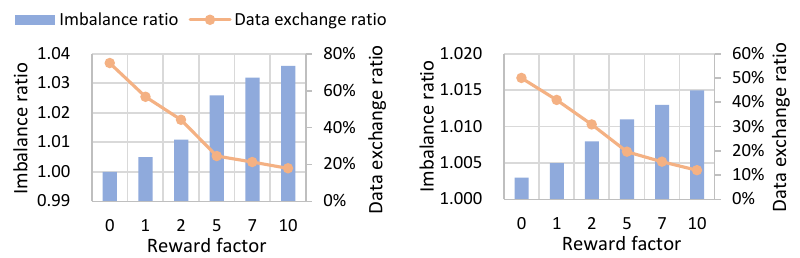}
    \vspace{-1.5em} % 也相应调整垂直间距
    \caption{Effect of reward factor $R_b$ choice. }
    \vspace{-1.0em}
    \label{fig:reward}
\end{figure}

% The overhead of \nickname comes from two aspects: the reordering of data during runtime and the redetermining of the partitioning plan for the dynamic sparse attention mechanism. For sequence-parallel models, each GPU initially holds the complete head dimension of query, key, and value, while holding $1/m$ in the sequence dimension, where n represents the number of devices. So for head-level partitioning, query, key, and value can be reordered before attention with a single line of code. Furthermore, the reordering of the key can overlap with the all-to-all communication of the query, and the reordering of the value can overlap with the all-to-all communication of the key, which reduces the overhead.  
Since head-level partitioning only involves intra-GPU reorderings, the overhead ratio in the end-to-end latency is less than 1\%. For further investigation, we quantify the effect of using different reusing thresholds $P_s$ on the \textit{sparse imbalance ratio} and the times of new plan generation, as shown in Fig.~\ref{fig:thr}. 
Although block-level partitioning needs inter-GPU data exchanges, the communication can be overlapped with the preceding computation, such as sorting and indexing. As a result, the overhead can be mitigated to within 5\% of the end-to-end latency. Moreover, we also quantify the effect of the reward factor $R_b$ choice, as depicted in Fig.~\ref{fig:reward}. A trade-off exists between the \textit{sparse imbalance ratio} and the associated overhead. We observe that the performance remains relatively stable when $R_b$ is set between 5 and 10 and $P_s$ between 1.03 and 1.10. Based on empirical observations, we select $R_b = 5$ and $P_s = 1.05$ as the default hyperparameters in our experiments.

% \textcolor{blue}{Tab.~\ref{tab:paro_attention} shows a typical breakdown timeline to indicate the latency of each part, especially the $index\_select$ and data exchange operations that constitute the overhead.}

To provide an intuitive understanding of the overhead, we present a typical end-to-end latency breakdown of Wan2.1 with PAROAttention in Tab.~\ref{tab:paro_attention}. The generation of the partitioning plan and parallel strategy incurs negligible overhead, as both are based on lightweight heuristics. The intra-GPU reorderings (\texttt{index\_select}) and inter-GPU data exchange introduce overheads on the order of seconds, which are further mitigated through overlap and remain minor compared to the overall speedup achieved in the attention.

\vspace{-0.5em}
\begin{table}[t]
\caption{A typical end-to-end latency breakdown of Wan2.1 and PAROAttention on eight A800 GPUs. O1,O2,..,O4 are overheads.}
\vspace{0.0em}
\label{tab:paro_attention}
\centering
% 使用 small 字体并略微缩小列间距，非常适合双栏环境
\small 
\setlength{\tabcolsep}{3pt} 
\begin{tabular}{
    >{\raggedright\arraybackslash}p{3cm} % 列1：固定3.2cm宽，左对齐且自动换行
    c                                      % 列2：居中紧凑
    >{\raggedright\arraybackslash}p{3.6cm} % 列3：固定4.4cm宽，左对齐且自动换行
}
\toprule
 & \textbf{Baseline} & \textbf{Ours} \\
\midrule
Attention & 104.5s & 63.5s (\textbf{1.65$\times$}) \\
\midrule

% O1
O1: Plan and strategy generation & -- & 0.013s (static sparsity) \\
\addlinespace % 行与行之间增加一点间距，使其不拥挤
\midrule
% O2
O2: Head-level \texttt{index\_select} & -- & 
1.334s ($Q$) + 1.342s ($K$) +  \newline 
1.518s ($V$) + 1.534s ($O$) = \newline 
5.728s (\textit{2.860s overlapped}) \\
\addlinespace
\midrule
% O3
O3: Block-level \texttt{index\_select} & -- & 
1.044s ($Q$) + 1.124s ($K$) \newline 
+ 1.116s ($V$) = 3.284s \\
\addlinespace
\midrule
% O4
O4: Inter-GPU data exchange & -- & 
\textit{1.024s ($Q$) + 1.002s ($K$)} + \newline 
0.956s ($V$) + 1.050s ($O$) = \newline 
4.032s (\textit{2.026s overlapped}) \\
\midrule

% Others
Others & 84.5s & 84.4s \\
\midrule
End-to-end & 189s & 156s (\textbf{1.21$\times$}) \\
\bottomrule
\end{tabular}
\vspace{-2.0em}
\end{table}

\section{Discussion}
In this section, we discuss whether \nickname can be generalized to other scenarios.

\textbf{Irregular Sparsity.} The irregular sparse pattern can be viewed as a form of block-wise sparsity with an extremely small block size (\textit{e.g.}, 1$\times$1), which is theoretically manageable by \nickname. However, as the block size decreases, the associated overhead becomes significant. Furthermore, smaller block sizes degrade the computation-to-memory-access ratio and introduce irregular memory access within the attention kernel. Consequently, current sparse attention methods \cite{zhang2025spargeattn,zhao2025paroattentionpatternawarereorderingefficient} for DiT inference adopt block sizes of 64/128, which balances generation quality with kernel efficiency. Compared to the sequence length exceeding 50k, block sizes of 64/128 remain sufficiently fine-grained to support diverse sparse patterns.

\textbf{DiT Training.} \nickname mainly focuses on DiT inference, while it is applicable for workload balancing in training. However, it requires additional considerations for the implementation complexity arising from backward computation and other parallelism strategies (\textit{e.g.}, tensor and data parallelism), as well as corresponding modifications to the latency prediction model for parallel strategy selection.

\textbf{LLMs.} The application to mainstream autoregressive LLMs presents several challenges. (1) Attention heads. LLMs tend to employ Grouped Query Attention (GQA) or Multi-head Latent Attention (MLA), which have fewer heads than the Multi-head Attention (MHA) commonly used in DiTs. A reduced head number enlarges workload imbalance under the head-level partitioning strategy of \nickname. (2) Limited applicability of SP. While DiTs feature relatively small model weights and inherently long sequence lengths, which are well-suited for SP, the advantages of SP for LLMs are confined to scenarios involving long contexts \cite{wu2024loongserveefficientlyservinglongcontext}.

% \section{Discussion}

\section{Conclusion}
In this paper, we introduce \nickname, a dual-balanced sequence parallelism method for accelerating visual generative tasks. \nickname applies tailored partitioning algorithms at the head level and block level to achieve a near-perfectly balanced workload across GPUs. Moreover, \nickname includes a parallel strategy selection mechanism that selects the optimal based on the sparse pattern at runtime. The techniques enable \nickname to outperform the existing sequence parallelism methods by 1.25$\times$ on average in end-to-end latency.

\section*{Acknowledgements}
We sincerely thank the anonymous shepherd and reviewers for their insightful suggestions. This work is supported by the Ministry of Industry and Information Technology of China (No. 2440STCZB2589), the National Natural Science Foundation of China (No. 62325405, U24B6015), Beijing Natural Science Foundation (No. L242018, L257010), Beijing Municipal Science and Technology Project (No. Z241100004224013), Beijing National Research Center for Information Science, Technology (BNRist), Beijing Innovation Center for Future Chips, and State Key Laboratory of Space Network and Communications.

% \textbf{Do not} include acknowledgements in the initial version of
% the paper submitted for blind review.

% If a paper is accepted, the final camera-ready version can (and
% probably should) include acknowledgements. In this case, please
% place such acknowledgements in an unnumbered section at the
% end of the paper. Typically, this will include thanks to reviewers
% who gave useful comments, to colleagues who contributed to the ideas,
% and to funding agencies and corporate sponsors that provided financial


\bibliography{example_paper}
\bibliographystyle{mlsys2025}

%%%%%%%%%%%%%%%%%%%%%%%%%%%%%%%%%%%%%%%%%%%%%%%%%%%%%%%%%%%%%%%%%%%%%%%%%%%%%%%
%%%%%%%%%%%%%%%%%%%%%%%%%%%%%%%%%%%%%%%%%%%%%%%%%%%%%%%%%%%%%%%%%%%%%%%%%%%%%%%
% SUPPLEMENTAL CONTENT AS APPENDIX AFTER REFERENCES
%%%%%%%%%%%%%%%%%%%%%%%%%%%%%%%%%%%%%%%%%%%%%%%%%%%%%%%%%%%%%%%%%%%%%%%%%%%%%%%
%%%%%%%%%%%%%%%%%%%%%%%%%%%%%%%%%%%%%%%%%%%%%%%%%%%%%%%%%%%%%%%%%%%%%%%%%%%%%%%
% \appendix
% \section{Please add supplemental material as appendix here}
% %
% Put anything that you might normally include after the references as an appendix here, {\it not in a separate supplementary file}. Upload your final camera-ready as a single pdf, including all appendices.

%%%%%%%%%%%%%%%%%%%%%%%%%%%%%%%%%%%%%%%%%%%%%%%%%%%%%%%%%%%%%%%%%%%%%%%%%%%%%%%
%%%%%%%%%%%%%%%%%%%%%%%%%%%%%%%%%%%%%%%%%%%%%%%%%%%%%%%%%%%%%%%%%%%%%%%%%%%%%%%

\appendix
\section{Artifact Appendix}

%%%%%%%%%%%%%%%%%%%%%%%%%%%%%%%%%%%%%%%%%%%%%%%%%%%%%%%%%%%%%%%%%%%%%
\subsection{Abstract}
This section is mainly a guideline to perform artifact evaluation for the paper \textit{db-SP: Accelerating Sparse Attention for Visual Generative Models with Dual-Balanced Sequence Parallelism}. The code of \nickname includes a dual-level balanced sequence parallel implementation and a sparsity-aware parallel strategy selection mechanism. The following artifacts demonstrate the functionality and efficiency of \nickname, including four experiments corresponding to the major claims in the paper: (1) The end-to-end speedup with \nickname. (2) The attention speedup against varying sparse imbalance ratios. (3) The effectiveness of the proposed parallel strategy selection method. (4) The negligible overhead against different hyperparameters.  

\subsection{Artifact check-list (meta-information)}

{\small
\begin{itemize}
  \item {\bf Algorithm: }Inference implementation of sparse attention on multi-GPU platforms.
  \item {\bf Program: }CUDA and Python code.
  \item {\bf Run-time environment: }Ubuntu 20.04 with CUDA SDK12.1 installed.
  \item {\bf Hardware: }Any Nvidia GPUs with computational capability $\geq$ 8.0 (Recommended GPU: NVIDIA A800 80GB or NVIDIA H800 80GB).
  \item {\bf How much disk space required (approximately)?: }30GB.
  \item {\bf How much time is needed to prepare workflow (approximately)?: }2 hours.
  \item {\bf How much time is needed to complete experiments (approximately)?: }3 hours.
  \item {\bf Publicly available?: }Yes.
\end{itemize}

%%%%%%%%%%%%%%%%%%%%%%%%%%%%%%%%%%%%%%%%%%%%%%%%%%%%%%%%%%%%%%%%%%%%%
\subsection{Description}
\subsubsection{How delivered}
The source code is available in the form of Github repository (\url{https://github.com/thu-nics/db-SP}). The sparse mask used for evaluation can be downloaded from Tsinghua Cloud (\url{https://cloud.tsinghua.edu.cn/d/458ccdacdf9c4edf8548/}).

\subsubsection{Hardware dependencies}
The implementation works on Intel x86 CPUs and Nvidia GPUs with the computational capability $\geq$ 8.0. For the convenience of evaluation, we provide a server equipped with Nvidia A800 GPUs. Please refer to \texttt{README.md} in the GitHub repository.  

\subsubsection{Software dependencies}
\begin{itemize}
    \item Python 3.8 (version not mandated)
    \item PyTorch 2.5.1
    \item NCCL 2.21.5
    \item Transformers $\geq$ 4.30.0
\end{itemize}

%%%%%%%%%%%%%%%%%%%%%%%%%%%%%%%%%%%%%%%%%%%%%%%%%%%%%%%%%%%%%%%%%%%%%
\subsection{Installation}

Please follow the \texttt{README.md} in the repository to install \nickname. The specific dependency is listed in \texttt{requirements.txt}, and a conda environment can be created via \texttt{environment.yaml}. 

%%%%%%%%%%%%%%%%%%%%%%%%%%%%%%%%%%%%%%%%%%%%%%%%%%%%%%%%%%%%%%%%%%%%%
\subsection{Experiment workflow}

Before the evaluation, please prepare the environments and download the models. On the provided server, we have already done the preparation steps. For evaluation, run the provided scripts according to \texttt{README.md} for each experiment. 
%%%%%%%%%%%%%%%%%%%%%%%%%%%%%%%%%%%%%%%%%%%%%%%%%%%%%%%%%%%%%%%%%%%%%
\subsection{Evaluation and expected result}

For each experiment, the results will be saved into the corresponding \texttt{.csv} files under the main directory. The latency and speedup numbers will be shown for comparison. 
\begin{itemize}
    \item End-to-end experiment shows about 1.25$\times$ speedup with \nickname on average.
    \item Attention experiment exhibits a close-to-linear growth trend as the sparse imbalance ratio increases.
    \item Parallel strategy selection enhances the overall performance by using different strategies for different layers. 
    \item Overhead can be negligible ($<$5\%) and provides an opportunity for a trade-off. 
\end{itemize}

%%%%%%%%%%%%%%%%%%%%%%%%%%%%%%%%%%%%%%%%%%%%%%%%%%%%
% When adding this appendix to your paper, 
% please remove below part
%%%%%%%%%%%%%%%%%%%%%%%%%%%%%%%%%%%%%%%%%%%%%%%%%%%%

\end{document}